\pdfoutput=1
\documentclass[a4paper,fleqn]{cas-sc}

\usepackage[numbers]{natbib}
\usepackage{amsmath,amssymb,amsfonts}

\usepackage{multirow}
\usepackage[normalem]{ulem}
\useunder{\uline}{\ul}{}
\usepackage{algorithmic}
\usepackage{booktabs}
\usepackage{graphicx}
\usepackage{threeparttable}
\usepackage[utf8]{inputenc}
\usepackage{textcomp}

\def\tsc#1{\csdef{#1}{\textsc{\lowercase{#1}}\xspace}}
\tsc{WGM}
\tsc{QE}
\tsc{EP}
\tsc{PMS}
\tsc{BEC}
\tsc{DE}

\begin{document}
\let\WriteBookmarks\relax
\def\floatpagepagefraction{1}
\def\textpagefraction{.001}

\shorttitle{A Survey on Legal Judgment Prediction}

\shortauthors{Junyun Cui et~al.}

\title [mode = title]{A Survey on Legal Judgment Prediction: Datasets, Metrics, Models and Challenges}                      



\author[1]{Junyun Cui}
\cormark[1]
\ead{junyun_cui@xaufe.edu.cn}
\affiliation[1]{organization={School of Information, Xi’an University of Finance and Economics},
    city={Xi'an},
    postcode={710100}, 
    state={Shaanxi},
    country={P. R. China}}

\author[2]{Xiaoyu Shen}

\affiliation[2]{organization={Saarland Informatics Campus},
    state={Saarbr\"ucken},
    country={Germany}}

\author[3]{Feiping Nie}
\author[3]{Zheng Wang}
\affiliation[3]{organization={School of Artificial Intelligence, Optics and Electronics (iOPEN), Northwestern Polytechnical University},
    city={Xi'an},
    postcode={710072}, 
    state={Shaanxi},
    country={P. R. China}}
\cortext[cor1]{Corresponding author}
\author[4]{Jinglong Wang}
\author[4]{Yulong Chen}
\affiliation[4]{organization={School of Law,  Xi’an University of Finance and Economics},
    city={Xi'an},
    postcode={710100}, 
    state={Shaanxi},
    country={P. R. China}}

\begin{abstract}
Legal judgment prediction (LJP) applies Natural Language Processing (NLP) techniques to predict judgment results based on fact descriptions automatically. Recently, large-scale public datasets and advances in NLP research have led to increasing interest in LJP. Despite a clear gap between machine and human performance, impressive results have been achieved in various benchmark datasets. In this paper, to address the current lack of comprehensive survey of existing LJP tasks, datasets, models and evaluations, (1) we analyze 31 LJP datasets in 6 languages, present their construction process and define a classification method of LJP with 3 different attributes; (2) we summarize 14 evaluation metrics under four categories for different outputs of LJP tasks; (3) we review 12 legal-domain pretrained models in 3 languages and highlight 3 major research directions for LJP; (4) we show the state-of-art results for 8 representative datasets from different court cases and discuss the open challenges.
This paper can provide up-to-date and comprehensive reviews to help readers understand the status of LJP. We hope to facilitate both NLP researchers and legal professionals for further joint efforts in this problem.
\end{abstract}



\maketitle
\section{Introduction}\label{sec:introduction}

Legal Judgment Prediction (LJP), which predicts the outcome of legal cases from the fact descriptions~\cite{lawlor1963computers}, plays a vital role in legal assistance systems and benefits both legal practitioners and normal citizens. In reality, this task is purely done by legal experts, who receive many years of specialized training before processing legal cases. LJP involves several steps like finding relevant law articles/history cases, defining the range of the charge, deciding the penalty term, etc. Each step is time-consuming and requires domain-specific solid background, incurring a great burden to the limited amounts of legal practitioners. In Louisiana, every attorney handles up to 50 cases per day, leaving only 1-5 minutes for case preparation~\cite{oppel2019one}. There are 332 thousand cases in progress in Brazil per day, considering only the financial domain~\cite{Brazil2014}. In India, 38 million pending cases cannot be handled on time until December 2021~\cite{NJDG2021backlog}. 
The massive imbalance between the need for legal assistance and the number of legal experts brings severe social problems. For example, low-income Americans reported 86\% of the civil legal issues received inadequate or no legal help ~\cite{legal2017justice}. This calls for the need for automatic LJP systems, which can significantly improve the working efficiency of legal experts since they do not need to process everything from scratch. Automated LJP systems can also offer real-time legal consult and enhance public access to justice. 

\begin{figure}
 \centering
  \includegraphics[width=6.2in]{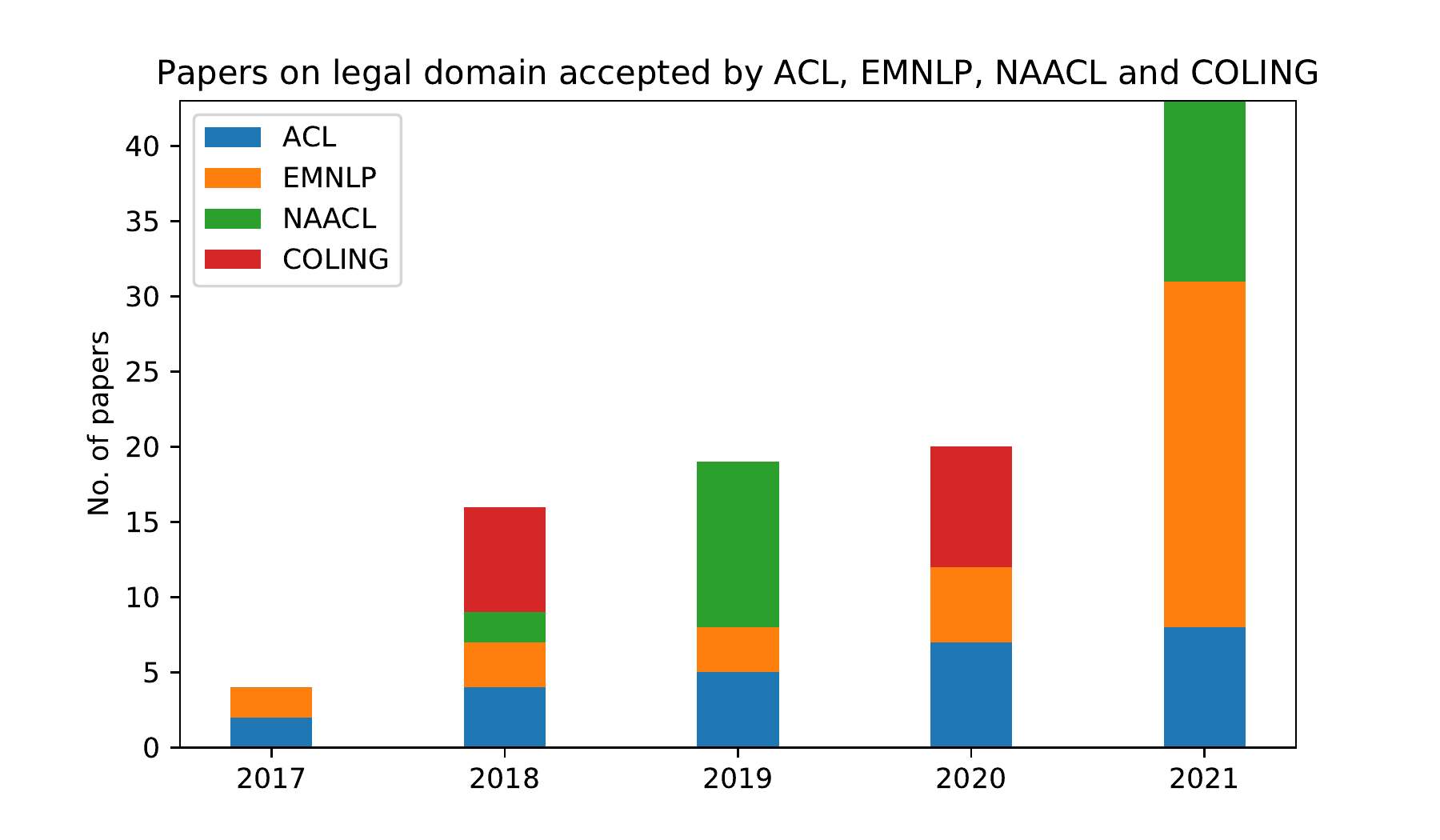}
  \caption{The number of legal-domain papers in major NLP conferences}\label{fig7}
\end{figure}

LJP is itself a long-standing task. As early as the 1950s, ~\cite{kort1957predicting,kort1960quantitative} have applied factor and linear regression analyses to predict decisions (pro or con) of the US Supreme Court cases depending on the 26 factual elements (patterns) with 14 training and 14 test "right to counsel" cases. Most LJP methods in the same period were based on rules or statistical methods~\cite{nagel1960weighting,ulmer1963quantitative,lawlor1963computers,keown1980mathematical}. However, these systems are sensitive to noise and cannot generalize to other law domains. Later on, researchers began to apply machine learning methods trained on a larger number of legal cases~\cite{liu2006exploring,katz2012quantitative,RN220,sulea2017predicting}.
Recently, the rapid advances of neural networks and large-scale pretrained language models based on the Transformer architecture further led to a considerable improvement in this area~\cite{2017Learning,shen2017estimation,vaswani2017attention,devlin-etal-2019-bert,brown2020language,ge2021learning,huang2021dependency}. As shown in Fig.~\ref{fig7}, the legal domain papers have proliferated in major NLP conferences. Among all these papers, around 65\% of them are relevant to the LJP task.

The availability of challenging benchmark datasets plays a crucial role in spurring innovation in LJP~\cite{kalamkar2021indian}. In the past few years, we have witnessed an explosion of public LJP benchmark datasets, such as CAIL2018~\cite{zhong2020does,xiao2018cail2018}, ECHR-CASES~\cite{zhong2020does,2019Neural}, SwissJudgmentPrediction~\cite{niklaus2021swiss}, JUSTICE~\cite{2021JUSTICE2021,chalkidis2021lexglue}. These datasets inspired a large number of LJP models, such as TopJudge~\cite{2018Legal}, MLCP-NLN~\cite{wei2019external}, MPBFN-WCA~\cite{2019Legal}, and LADAN~\cite{2020Distinguish}. Impressive results have been achieved in various benchmark datasets, despite a clear gap between the machine and human performance~\cite{xiao2018cail2018}. However, most researchers focus on a few popular LJP datasets, while most other LJP datasets are not widely known and studied by the community. In addition, there is also a need for systematic categorization/classification of LJP subtasks. For example, LJP tasks are generally divided into three subtasks: the decision of applicable law articles, charges, and terms of penalty~\cite{xiao2018cail2018,2018Legal,2019Legal,2020Distinguish}, but this classification method is not thorough, only applies to limited legal systems and domains~\cite{Hai2018Interpretable,wu-etal-2020-de}. A comprehensive survey of existing LJP tasks, datasets, evaluation metrics, and models is strongly needed to promote the future development of LJP.

This paper aims to address this gap. There have been a few surveys in the LJP domain. For example, experimental results for symbolic and embedding-based methods and possible future directions have been provided in~\cite{zhong2020does}, but their effects are based on the single benchmark dataset CAIL2018. The task definition and evaluation metrics have been mentioned in~\cite{kalamkar2021indian}, but they focus only on Indian Legal NLP benchmarks and do not cover advances made in other benchmarks. To the best of our knowledge, our survey is the first work that provides a comprehensive survey of the LJP task. It provides a thorough overview of 31 publicly available LJP datasets in 6 languages and introduces the pros and cons of popular state-of-the-art models. 

In section~\ref{Tasks Formulations}, we introduce the basis of LJP tasks. We first compare the litigation procedure differences between the common-law and the civil-law system. Next, we define some notations of and analyze the critical dimensions of the existing 31 LJP datasets. We summarized three attributes of LJP: the type of LJP tasks, the type of legal systems, and the type of law domains. Based on the outcomes of LJP tasks, we propose a thorough taxonomy to classify existing LJP datasets. 

Section~\ref{Datasets} presents 31 LJP datasets in 6 languages and 6 pre-training datasets in 3 languages. In section~\ref{Evaluation Metrics}, evaluation metrics are introduced. We begin by analyzing the four categories for evaluating the outputs from LJP tasks. Then we discuss the computing methods of 14 evaluation metrics of LJP tasks among three LJP task types. Finally, we present the usages of different evaluation metrics in the existing 31 LJP tasks.

Section~\ref{methods} reviews 12 legal-domain pretrained models in 3 languages and highlights 3 other major research directions, including multi-task learning, interpretable learning, and few-shot learning models. Multi-task learning methods concentrate on utilizing the dependent information among the multi-subtask in LJP. Interpretable learning methods aim to enhance the interpretability of LJP models and few-shot learning targets at improving the data efficiency on low-frequency (i.e., few-shot) judgment results.  In section~\ref{result}, we share the state-of-art results for eight representative LJP datasets from different court cases and the observations of these results for explored research.

Finally, in section~\ref{Discussions}, we summarize three prime challenges for existing datasets and methods in LJP: (1) LJP when there is inadequate information from facts. In the trial practice, sometimes an event appears to fit within the statute's language, but it seems absurd to think that the legislature intended to make the disputed activity a crime. At other times, the defendant has done something similar to the activities included in the statutory language, but it seems like a stretch to say that the words fit the event in question. LJP, when there is inadequate information from the determination of facts by the jury, will significantly prevent civil and criminal errors. Predicting the judgment results reducing the universe of evidence from real courtrooms, especially when the proof rules are inadequate or the text is ambiguous, would be a proactive issue in the future. (2) Complex legal logical reasoning. Current NLP models are good at exploring the statistics in the dataset to achieve good accuracy but  cannot still reason over complex evidence, deduce from facts and find the logical dependence among factors. (3) Interpretability. LJP is a critical task, and any tiny error from the system might affect judicial fairness. Without good interpretability, humans can hardly trust the output from the automatic LJP system proactive.

\section{Tasks Formulations}\label{Tasks Formulations}
\subsection{Background}\label{background}
\begin{figure}
 \centering
  \includegraphics[width=6.2in]{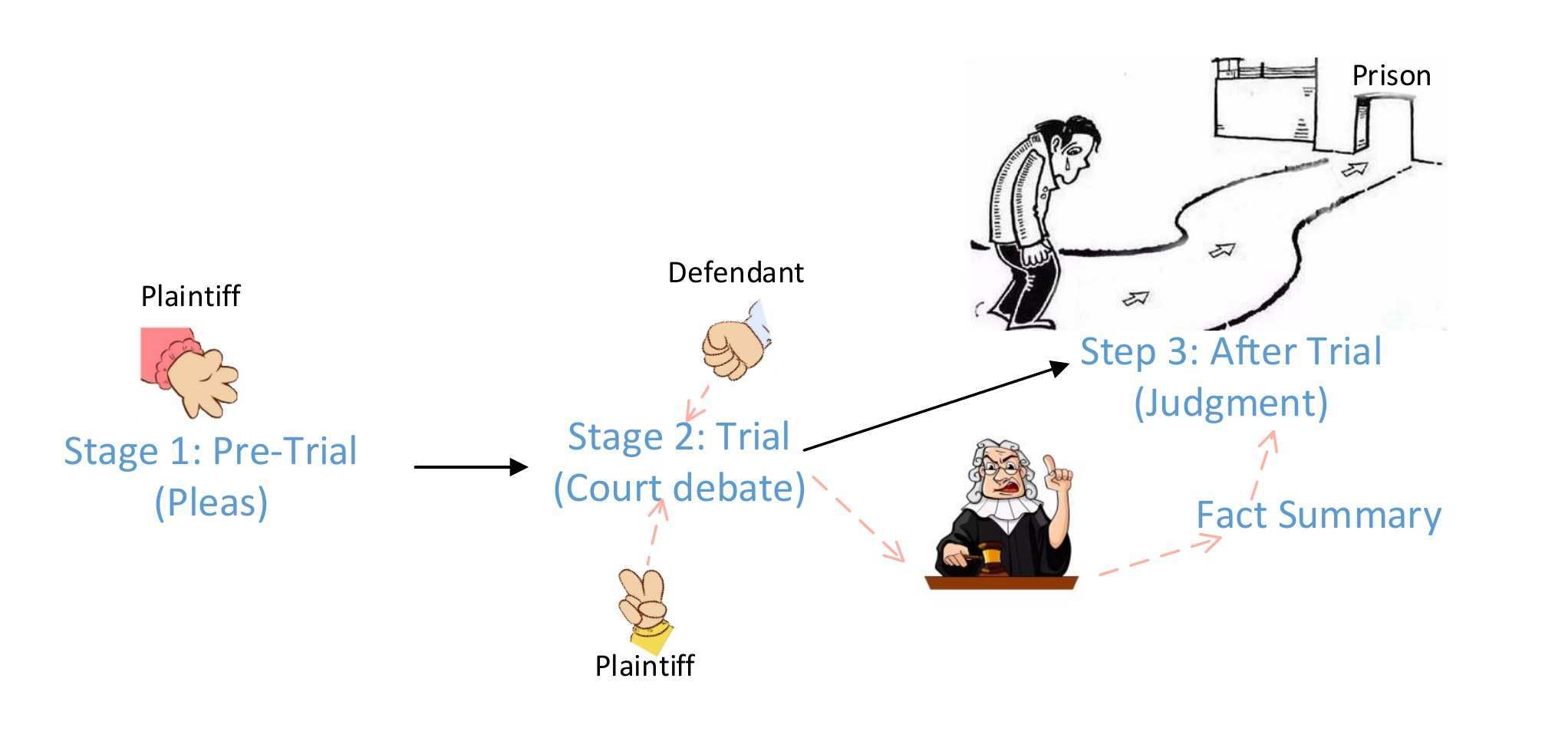}
  \caption{The procedure of court judgment. Firstly, the plaintiff submits his pleas. Secondly, fact verdict is made based on the court debate held between the plaintiff and defendant. Finally, the judge make decisions based on the fact verdict.}\label{fig:cut}
\end{figure}
As is well known, continental civil-law of those countries (e.g., France, Germany, Switzerland, Belgium, and the Netherlands in European continental, China, Thailand, and Vietnam in Asiatic countries, and Scandinavian countries and Soviet countries) and the common-law of the Anglo-Saxon countries(e.g., the United Kingdom, the United States, Canada, Australia, New Zealand, and India) are two significant legal systems. Both these legal systems have the same purpose of regulating and harmonizing the human activity within their respective societies, and in each society, its legal system is also influenced by its culture, epistemology, and civilization as well as of the history and the life of its people, such as the civil-law and the common-law have been functioning together in countries like Japan and Italy and places like Louisiana, Quebec, Scotland and South Africa. However, it must be recognized that there are of course significant differences between common-law and civil-law. As Roman-law writers had noted that the Roman law of the classical period, the first two centuries A.D. when it reached its highest point of technical development, is in many respects closer in character to the common-law than it is to modern civil-law systems that are derived from Roman law ~\cite{stein1991roman}.  

As shown in Fig. \ref{fig:cut}, a judicial procedure in a real court setting abides by the no claim no trial principle, whereas the court's claimant's pleadings and court's decisions are two essential components of judicial procedure to protect the legitimate interest of parties. Specifically, a litigation procedure often experiences three stages: pre-trial claim collection stage (plaintiff or petitioner submit written materials appealing its case to the court), trial court debate stage (Followed that the court decides to hear that case, the parties containing plaintiff, defendant, witness, lawyer debate before the court focusing on the fact details), and after trial judge sentence stage(judge generates verdict including judgment).

Around the litigation procedure above, the differences between civil-law and common-law system are further concluded:

\begin{itemize}
    \item Pre-trial claim collection stage. The principle of prosecuting a case (legality in civil-law vs. opportunity in common-law)~\cite{ligeti2019place}: Under the legality principle, if there is sufficient evidence at the close of the investigation, the prosecutor is in principle duty-bound to press charges and cannot dismiss a case. The opportunity principle, on the contrary, entrusts the prosecutor with the
discretionary choice of whether or not to prosecute a case despite sufficient evidence 
to charge. Based on the opportunity principle, the portion of federal civil cases resolved by trial fell from
11.5 percent in 1962 to 1.8 percent in 2002, and a similar decline in both the percentage and the
absolute number of trials in federal criminal cases~\cite{galanter2004vanishing}, whereas the accused’s 
fate~\cite{mckillop2003position} is determined by the trial court in common-law rather than the conclusions drawn from the investigation in civil-law. 
    \item Trial court debate stage. Two differences are concluded: (1) Formality of evidence (Public oral evidence in common-law vs. private written proof in civil-law)~\cite{stein1991roman}: In the common-law, the witnesses must give their evidence in public before the court while the civil-law has a preference for written proof in private before a judge who questions him from a neutral standpoint. (2) Methods of facts verdict (Voting jury trial in common-law vs. collegial reasoning trial in civil-law)~\cite{dainow1966civil}: The jury in common-law based on its voting values never had to justify its fact verdict with reasons whereas the judge in civil-law must give the reasons for its verdict based on given facts, the established statutory law articles and common sense.
    \item After trial judge sentence stage (precedent in common-law vs. legislation in civil-law)~\cite{dainow1966civil}. The civil-law judges search the legislation 
for the controlling principle and the rules which govern the subject; 
this principle or rule is then applied or interpreted according to the 
particular facts of the dispute case. On the contrary, if a statute
is involved and the text is clear, common-law judges abide by its provisions. However, 
if doubt or ambiguity can avoid the statute's applicability, common-law judges search in the previous decisions for a similar case and are guided accordingly, this precedent is then applied or interpreted according to the determination of facts by the jury, the summarized evidence and the relevant rules of law.
   \end{itemize}

\begin{figure*}
 \centering
  \includegraphics[width=6.4in]{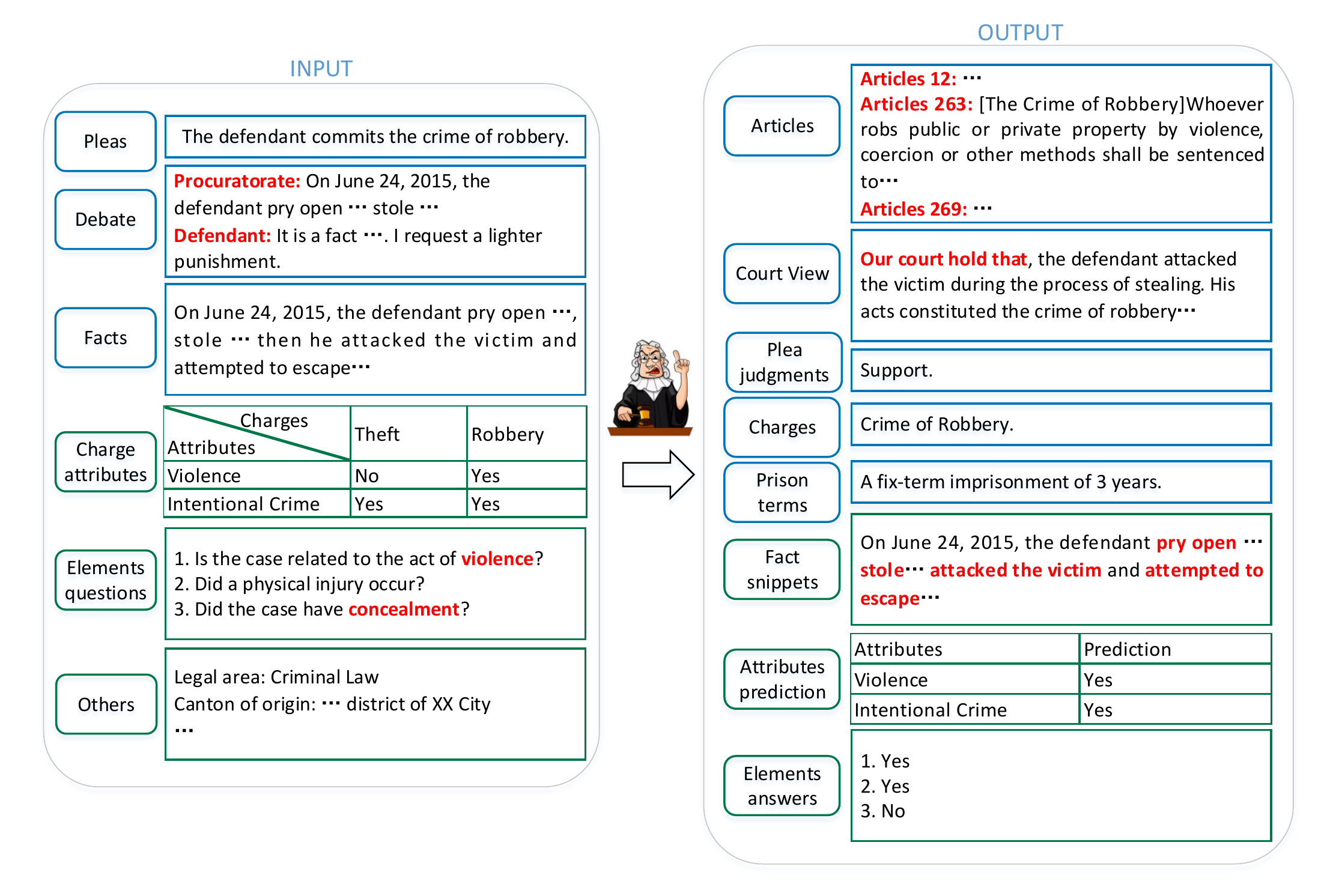}
  \caption{An outcomes-based judicial framework example interpretation of notations introduced by a law case life-cycle in a real court setting. Within this framework, the LJP tasks are divided into two categories: main LJP tasks (blue) and auxiliary LJP tasks (green).}\label{fig:example}
\end{figure*}
Based on the background of LJP tasks, some notations of LJP tasks for a court case instance in Fig.~\ref{fig:example} are given, as shown in Table~\ref{table: notation}.

\begin{table}
\centering
\caption{Notations definition for a court case.}
\label{table: notation}
\begin{tabular}{|l|l|}
\hline
\textbf{Notation}     & \textbf{Description}                                                                                                                                                                       \\ \hline
Pleas                 & The sentence narratives from the plaintiff for the target dispute.                                                                                                                         \\ \hline
Plea Judgments        & \begin{tabular}[c]{@{}l@{}}The response sentence from the judge on the pleas with labels like reject,\\  partially support and support.\end{tabular}                                       \\ \hline
Debate                & \begin{tabular}[c]{@{}l@{}}The diachronic statement of fact detail from the plaintiff, defendant, witness, \\ lawyer and judge in the real court on the target dispute.\end{tabular}       \\ \hline
Facts                 & The admissible facts summarized by the judge by employing evidence                                                                                                                         \\ \hline
Articles              & The applied law articles of a Law case.                                                                                                                                                    \\ \hline
Charges               & The specific statement of what crime the party is accused contained in the criminal plea.                                                                                                                                     \\ \hline
Prison terms          & The penalty terms for corresponding charges in a criminal case.                                                                                                                            \\ \hline
Court View            & \begin{tabular}[c]{@{}l@{}}The explanation written by judges to interpret the judgment decision \\ for certain law case.\end{tabular}                                                      \\ \hline
Facts snippets        & The extracted decisive phrases or sentences from facts.                                                                                                                                    \\ \hline
Charge attributes     & The predefined attribute knowledge for discriminating confusing charges.                                                                                                                   \\ \hline
Attributes prediction & \begin{tabular}[c]{@{}l@{}}A binary classification task used to predict the charge with the attribute or not \\ according to the input facts.\end{tabular}                                 \\ \hline
Elements questions    & The predefined questions for extracting judgment elements.                                                                                                                                 \\ \hline
Elements answers      & The answers for element questions.                                                                                                                                                         \\ \hline

Others                & \begin{tabular}[c]{@{}l@{}}The additional metadata that judges can obtain from a case except the items described above, \\ such as the date, the legal area, the canton of origin per case.\end{tabular} \\ \hline
\end{tabular}
\end{table}

\begin{figure}
 \centering
  \includegraphics[width=5.8in]{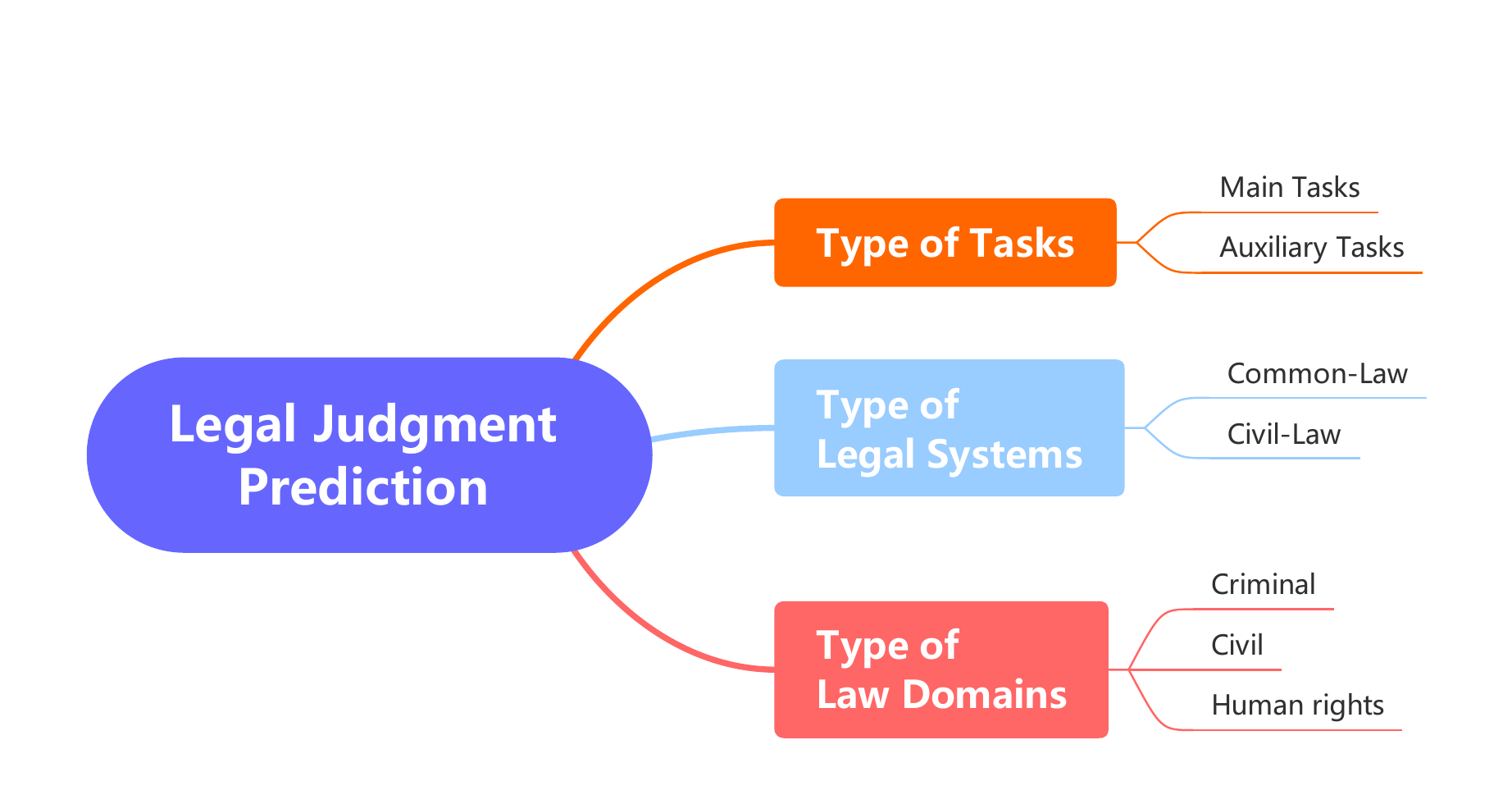}
  \caption{Proposed classification method of legal judgment prediction tasks.}\label{fig:classification}
\end{figure}
\subsection{Taxonomy}
To have a better understanding of LJP tasks, in this section, we analyze existing LJP tasks from the key dimensions of 31 LJP datasets as shown in Table \ref{table00} based on notations and terminologies in section \ref{background}. In many research ~\cite{xiao2018cail2018,2018Legal,2019Legal,2020Distinguish}, LJP tasks are divided into three subtasks: 
the decision of applicable law articles, charges, and penalty terms. However, as shown in Table~\ref{table00}, this classification method has limited coverage of LJP tasks, such as neglecting pro or con decisions. Hence, as shown in Fig.~\ref{fig:classification}, we propose a new classification method of LJP tasks with summarized three attributes of LJP tasks, including the type of tasks, the type of legal systems, and the type of law domains. Each of these attributes can be divided into several different categories. These categories are: 

\begin{itemize}
    \item Type of tasks. Two cases are conducted: (1) main tasks: Judgements results are determined by these tasks' outcomes, such as Article recommendation, Charge prediction, Prison term prediction, Court View generation, and Plea judgment task in Table~\ref{table00} and Fig.~\ref{table00}. (2) auxiliary tasks: Judgements results are latently improved by these tasks' outcomes, such as Fact snippets extraction, attribute prediction of the confusing charges and elements answers.
    \item Type of legal systems. Two main legal systems are considered, including the common-law and civil-law systems. Specifically, for the common-law system: (1) Fact verdicts: Verdicts of Fact gives out the voting values for their Court's charges. (2) Sentence: The trial judge provides the Court's results, including law article recommendation, precedent, prison terms, and Plea judgments. For the civil-law system: (1) Justify the judgments: The judge gives the reasons for its judgment results based on given facts, the established statutory law articles, and common sense. (2) Judgment results: The trial judge provides the Court's results, including charges, prison terms, and Plea judgments. 
    \item Type of law domains. According to online legal community users' demand and the case classification that the Supreme People's Court of the People's Republic of China provides in the Chinese Judgments Online, three prominent cases are conducted: (1)criminal: Articles, Charges, Prison terms, and Plea judgments. (2)civil: Articles, Obligation, Penalty terms, and Plea judgments. (3)administrative: Articles, Penalty terms, and Plea judgments.  
   \end{itemize}

\subsubsection{Types of LJP Tasks}
\begin{table}
\centering
\caption{The key dimensions for the format of the existing LJP tasks.}
\label{table00}
\setlength{\tabcolsep}{0.01pt}
\begin{tabular}{|c|c|c|c|c|c|c|c|c|c|}
\hline
\multirow{2}{*}{Dataset}                                                 & \multirow{2}{*}{Language} & \multicolumn{3}{c|}{\multirow{2}{*}{Input}}                                        & \multicolumn{4}{c|}{Output}                                                                                              & \multirow{2}{*}{Download} \\ \cline{6-9}
                                                                         &                           & \multicolumn{3}{c|}{}                                                              & \multicolumn{2}{c|}{Rationale}                                & \multicolumn{2}{c|}{Decision}                            &                           \\ \hline
\multirow{3}{*}{SwissJudgmentPrediction ~\cite{niklaus2021swiss}} & German                    & \multicolumn{1}{c|}{\multirow{4}{*}{Pleas$^{\rm r}$}} & \multicolumn{2}{c|}{\multirow{3}{*}{Facts$^{\rm r}$}}                           & \multicolumn{2}{c|}{\multirow{4}{*}{-}}                                                                 & \multicolumn{2}{c|}{\multirow{4}{*}{Plea judgments$^{\rm r}$}}                          & \multirow{3}{*}{\href{https://zenodo.org/record/5529712}{Link}}     \\ \cline{2-2}
                                                                         & Italian                   & \multicolumn{1}{c|}{}                       & \multicolumn{2}{c|}{}                                                 & \multicolumn{2}{c|}{}                                                                                   & \multicolumn{2}{c|}{}                                                         &                           \\ \cline{2-2}
                                                                         & \multirow{2}{*}{French}   & \multicolumn{1}{c|}{}                       & \multicolumn{2}{c|}{}                                                 & \multicolumn{2}{c|}{}                                                                                   & \multicolumn{2}{c|}{}                                                         &                           \\ \cline{1-1} \cline{4-5} \cline{10-10} 
Sulea et al.~\cite{sulea2017predicting}                 &                           & \multicolumn{1}{c|}{}                       & \multicolumn{1}{c|}{Facts$^{\rm r}$}                  & Others$^{\rm r}$                  & \multicolumn{2}{c|}{}                                                                                   & \multicolumn{2}{c|}{}                                                         & -                         \\ \hline
TSCC~\cite{kowsrihawat2018predicting}                                                                    & Thai                      & \multicolumn{3}{c|}{Facts$^{\rm a}$}                                                                                          & \multicolumn{2}{c|}{Articles$^{\rm a}$}                                                                           & \multicolumn{2}{c|}{-}                                                        & \href{https://github.com/KevinMercury/tscc-dataset}{Link}                      \\ \hline
JUSTICE~\cite{2021JUSTICE2021}                                                                  & \multirow{7}{*}{English}  & \multicolumn{1}{c|}{\multirow{3}{*}{Pleas$^{\rm r}$}} & \multicolumn{2}{c|}{Facts$^{\rm r}$}                                            & \multicolumn{2}{c|}{\multirow{2}{*}{-}}                                                                 & \multicolumn{2}{c|}{\multirow{3}{*}{Plea judgments$^{\rm r}$}}                          & -                         \\ \cline{1-1} \cline{4-5} \cline{10-10} 
BStricks\_LDC~\cite{strickson2020legal}              &                           & \multicolumn{1}{c|}{}                       & \multicolumn{1}{c|}{\multirow{2}{*}{Facts$^{\rm r}$}} & \multirow{2}{*}{Others$^{\rm r}$} & \multicolumn{2}{c|}{}                                                                                   & \multicolumn{2}{c|}{}                                                         & \href{https://github.com/BStricks/legal_document_classifier_V2}{Link}                      \\ \cline{1-1} \cline{6-7} \cline{10-10} 
ILDC~\cite{malik2021ildc}                                                                     &                           & \multicolumn{1}{c|}{}                       & \multicolumn{1}{c|}{}                       &                         & \multicolumn{2}{c|}{\multirow{2}{*}{Fact snippets$^{\rm r}$}}                                                     & \multicolumn{2}{c|}{}                                                         & \href{https://github.com/Exploration-Lab/
CJPE}{Link}                      \\ \cline{1-1} \cline{3-5} \cline{8-10} 
ACI~\cite{paul-etal-2020-automatic}              &                           & \multicolumn{3}{c|}{Facts$^{\rm a,r}$}                                                                             & \multicolumn{2}{c|}{}                                                                                   & \multicolumn{2}{c|}{Charges$^{\rm a}$}                                                  & \href{https://github.com/Law-AI/automatic-charge-identification}{Link}                       \\ \cline{1-1} \cline{3-10} 
echr~\cite{RN220}                                                   &                           & \multicolumn{3}{c|}{\multirow{4}{*}{Facts$^{\rm r}$}}                                                            & \multicolumn{2}{c|}{\multirow{4}{*}{Articles$^{\rm r}$}}                                                          & \multicolumn{2}{c|}{\multirow{4}{*}{-}}                                       & \href{https://figshare.com/s/6f7d9e7c375ff0822564}{Link}                      \\ \cline{1-1} \cline{10-10} 
ECHR-CASES~\cite{2019Neural}                                                 &                           & \multicolumn{3}{c|}{}                                                                                  & \multicolumn{2}{c|}{}                                                                                   & \multicolumn{2}{c|}{}                                                         & \href{https://archive.org/details/ECHR-ACL2019}{Link}                       \\ \cline{1-1} \cline{10-10} 
ECtHR\_crystal\_ball~\cite{2018Medvedeva}                                                 &                           & \multicolumn{3}{c|}{}                                                                                  & \multicolumn{2}{c|}{}                                                                                   & \multicolumn{2}{c|}{}                                                         & \href{https://github.com/masha-medvedeva/ECtHR\_crystal\_ball}{Link}                       \\ \cline{1-2} \cline{10-10} 
DPAM~\cite{RN224}                                                      & \multirow{21}{*}{Chinese} & \multicolumn{3}{c|}{}                                                                                  & \multicolumn{2}{c|}{}                                                                                   & \multicolumn{2}{c|}{}                                                         & \href{https://drive.google.com/open?id=1TCcTzte2cQ2wxWw-kXsZdUApCsNejdn8}{Link}                      \\ \cline{1-1} \cline{3-10} 
MLMN~\cite{2021Learning}                                                        &                           & \multicolumn{3}{c|}{Facts$^{\rm a}$}                                                                             & \multicolumn{2}{c|}{Articles$^{\rm a}$}                                                                           & \multicolumn{2}{c|}{Prison terms$^{\rm r}$}                                             & \href{https://github.com/gjdnju/MLMN}{Link}                      \\ \cline{1-1} \cline{3-10} 
CAIL2018~\cite{xiao2018cail2018}                                                                &                           & \multicolumn{3}{c|}{\multirow{9}{*}{Facts$^{\rm r}$}}                                                            & \multicolumn{2}{c|}{\multirow{6}{*}{Articles$^{\rm r}$}}                                                          & \multicolumn{1}{c|}{\multirow{5}{*}{Charges$^{\rm r}$}} & \multirow{5}{*}{Prison terms$^{\rm r}$} & \href{http://cail.cipsc.org.cn/index.html}{Link}                       \\ \cline{1-1} \cline{10-10} 
TOPJUDGE-CJO~\cite{2018Legal}                                                        &                           & \multicolumn{3}{c|}{}                                                                                  & \multicolumn{2}{c|}{}                                                                                   & \multicolumn{1}{c|}{}                         &                               & \multirow{3}{*}{-}        \\ \cline{1-1}
TOPJUDGE-PKU~\cite{2018Legal}                                                         &                           & \multicolumn{3}{c|}{}                                                                                  & \multicolumn{2}{c|}{}                                                                                   & \multicolumn{1}{c|}{}                         &                               &                           \\ \cline{1-1}
TOPJUDGE-CAIL~\cite{2018Legal}                                                        &                           & \multicolumn{3}{c|}{}                                                                                  & \multicolumn{2}{c|}{}                                                                                   & \multicolumn{1}{c|}{}                         &                               &                           \\ \cline{1-1} \cline{10-10} 
CAIL-Long~\cite{xiao2021lawformer}                                                               &                           & \multicolumn{3}{c|}{}                                                                                  & \multicolumn{2}{c|}{}                                                                                   & \multicolumn{1}{c|}{}                         &                               &  \href{https://github.com/thunlp/LegalPLMs}{Link}                      \\ \cline{1-1} \cline{8-10} 
FLA~\cite{2017Learning}                                                        &                           & \multicolumn{3}{c|}{}                                                                                  & \multicolumn{2}{c|}{}                                                                                   & \multicolumn{2}{c|}{\multirow{3}{*}{Charges$^{\rm r}$}}                                 & \multirow{3}{*}{-}        \\ \cline{1-1} \cline{6-7}
RACP~\cite{2018Interpretable}                                                    &                           & \multicolumn{3}{c|}{}                                                                                  & \multicolumn{2}{c|}{Fact snippets$^{\rm a}$}                                                                      & \multicolumn{2}{c|}{}                                                         &                           \\ \cline{1-1} \cline{6-7}
MAMD~\cite{RN200}                                                      &                           & \multicolumn{3}{c|}{}                                                                                  & \multicolumn{2}{c|}{-}                                                                                  & \multicolumn{2}{c|}{}                                                         &                           \\ \cline{1-1} \cline{6-10} 
Court-View-Gen~\cite{Hai2018Interpretable}                                                        &                           & \multicolumn{3}{c|}{}                                                                                  & \multicolumn{4}{c|}{\multirow{2}{*}{Court View$^{\rm r}$}}                                                                                                                                        & \href{https://github.com/oceanypt/Court-View-Gen}{Link}                      \\ \cline{1-1} \cline{3-5} \cline{10-10} 
AC-NLG~\cite{wu-etal-2020-de}                         &                           & \multicolumn{1}{c|}{\multirow{8}{*}{Facts$^{\rm r}$}} & \multicolumn{2}{c|}{Pleas$^{\rm r}$}                               & \multicolumn{4}{c|}{}                                                                                                                                                                   & \href{https://github.com/wuyiquan/AC-NLG}{Link}                      \\ \cline{1-1} \cline{4-10} 
CPTP~\cite{2019Charge}                                                      &                           & \multicolumn{1}{c|}{}                       & \multicolumn{2}{c|}{Charges$^{\rm r}$}                             & \multicolumn{2}{c|}{-}                                                                                  & \multicolumn{2}{c|}{Prison terms$^{\rm r}$}                                             & \href{https://github.com/huajiechen/CPTP}{Link}                      \\ \cline{1-1} \cline{4-10} 
Criminal-S~\cite{2018few-shot}                                                              &                           & \multicolumn{1}{c|}{}                       & \multicolumn{2}{c|}{\multirow{3}{*}{Charge attributes$^{\rm a}$}}  & \multicolumn{2}{c|}{\multirow{3}{*}{Attributes prediction$^{\rm r}$}}                                             & \multicolumn{2}{c|}{\multirow{6}{*}{Charges$^{\rm r}$}}                                 & \href{https://github.com/thunlp/attribute\_charge}{Link}                      \\ \cline{1-1} \cline{10-10} 
Criminal-M~\cite{2018few-shot}                                                              &                           & \multicolumn{1}{c|}{}                       & \multicolumn{2}{c|}{}                                    & \multicolumn{2}{c|}{}                                                                                   & \multicolumn{2}{c|}{}                                                         & \href{https://github.com/thunlp/attribute\_charge}{Link}                      \\ \cline{1-1} \cline{10-10} 
Criminal-L~\cite{2018few-shot}                                                              &                           & \multicolumn{1}{c|}{}                       & \multicolumn{2}{c|}{}                                    & \multicolumn{2}{c|}{}                                                                                   & \multicolumn{2}{c|}{}                                                         & \href{https://github.com/thunlp/attribute\_charge}{Link}                      \\ \cline{1-1} \cline{4-7} \cline{10-10} 
QAjudge-CJO~\cite{2020Iteratively}                                                       &                           & \multicolumn{1}{c|}{}                       & \multicolumn{2}{c|}{\multirow{3}{*}{Elements questions$^{\rm a}$}} & \multicolumn{1}{c|}{\multirow{3}{*}{Articles$^{\rm r}$}} & \multicolumn{1}{c|}{\multirow{3}{*}{Elements answers$^{\rm r}$}} & \multicolumn{2}{c|}{}                                                         & \multirow{4}{*}{-}        \\ \cline{1-1}
QAjudge-PKU~\cite{2020Iteratively}                                                        &                           & \multicolumn{1}{c|}{}                       & \multicolumn{2}{c|}{}                                    & \multicolumn{1}{c|}{}                          & \multicolumn{1}{c|}{}                                  & \multicolumn{2}{c|}{}                                                         &                           \\ \cline{1-1}
QAjudge-CAIL~\cite{2020Iteratively}                                                       &                           & \multicolumn{1}{c|}{}                       & \multicolumn{2}{c|}{}                                    & \multicolumn{1}{c|}{}                          & \multicolumn{1}{c|}{}                                  & \multicolumn{2}{c|}{}                                                         &                           \\ \cline{1-1} \cline{3-9}
Auto-Judge~\cite{2018Automatic}                                                     &                           & \multicolumn{1}{c|}{\multirow{2}{*}{Pleas$^{\rm r}$}} & \multicolumn{1}{c|}{Facts$^{\rm r}$}           & Articles$^{\rm r}$          & \multicolumn{2}{c|}{\multirow{2}{*}{-}}                                                                 & \multicolumn{2}{c|}{Plea judgments$^{\rm r}$}                                           &                           \\ \cline{1-1} \cline{4-5} \cline{8-10} 
LJP-MSJudge~\cite{RN202}                                                        &                           & \multicolumn{1}{c|}{}                       & \multicolumn{2}{c|}{Debate data$^{\rm r}$}                         & \multicolumn{2}{c|}{}                                                                                   & \multicolumn{2}{c|}{Plea judgments$^{\rm a}$}                                           & \href{https://github.com/mly-nlp/LJP-MSJudge}{Link}                       \\ \hline
\end{tabular}
\begin{tablenotes}
			\footnotesize
			\item[$^{\rm r}$]Extracted by machine.
\end{tablenotes}
\begin{tablenotes}
			\footnotesize
			\item[$^{\rm a}$]Annotated by legal experts.
\end{tablenotes}
\end{table}

\begin{figure}
 \centering
  \includegraphics[width=6.2in]{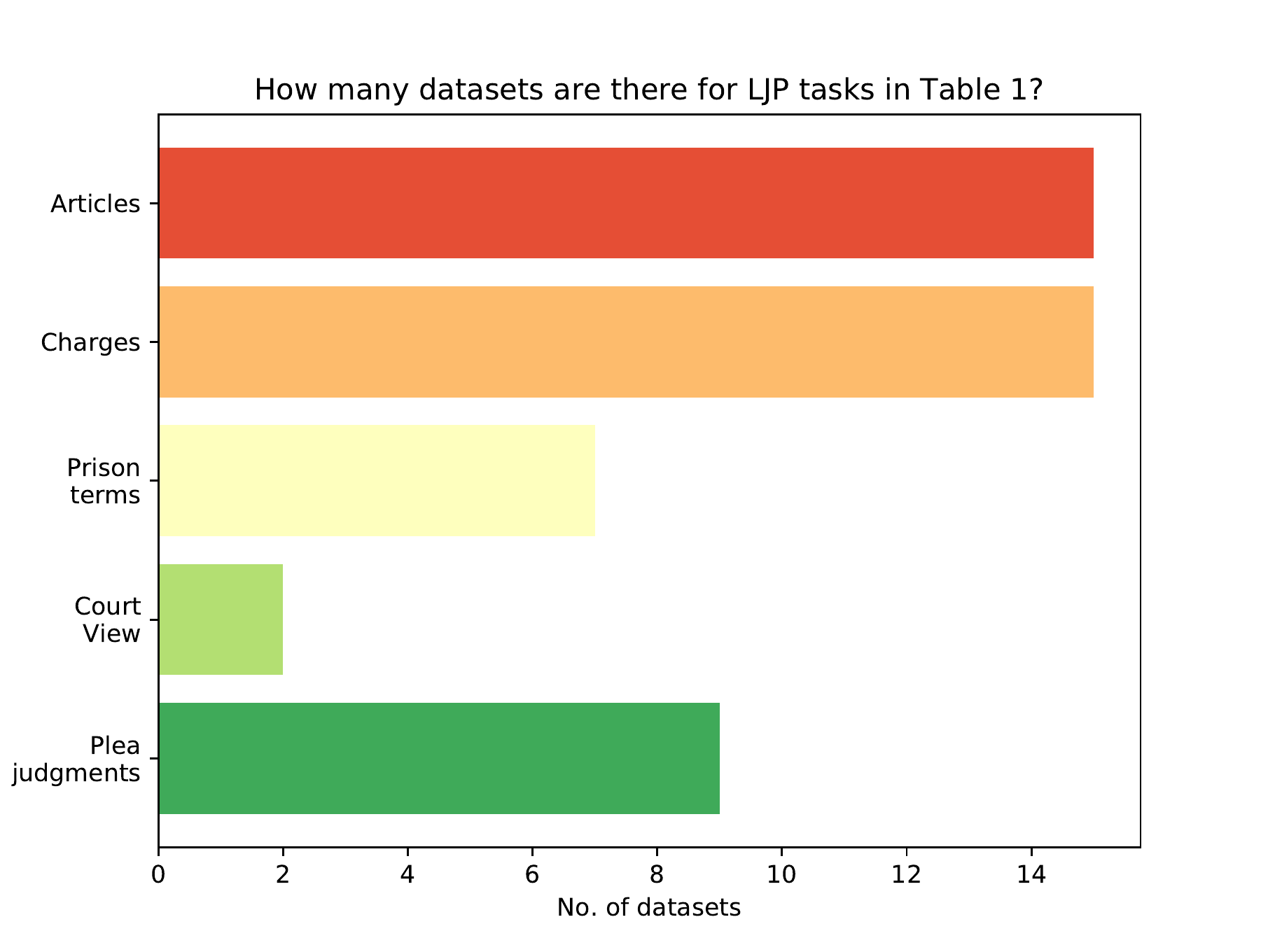}
  \caption{The dataset density distribution for specific LJP task. }\label{fig:distribution}
\end{figure}
 
 And then, under the proposed outcome-based taxonomy, existing LJP main tasks are briefly concluded as follows:

\textbf{Article Recommendation Task}. As seen in Table \ref{table00} and Fig.~\ref{fig:distribution}, the hottest LJP tasks are article recommendation and charge prediction. Furthermore, these articles recommend tasks are classified into three main parts:
many-vs-one, many-vs-many, one-vs-many according to the corresponding relationship between facts and their relevant articles as follows:

\begin{itemize}
    \item Many-vs-one. Two cases are conducted: (1) Binary Violation. echr~\cite{RN220}, ECHR-CASES~\cite{2019Neural}, and ECtHR\_crystal\_ball~\cite{2018Medvedeva} are built to determine whether any particular law article was violated or not for given case facts. (2)Multi-label Violation. ECHR-CASES~\cite{2019Neural}, and CAIL2018 are used to predict which specific article has been violated or directly applicable to the given case.
    \item Many-vs-many. DPAM~\cite{RN224}, FLA~\cite{2017Learning}, CAIL-Long found optimal article set for each case.
    \item One-vs-many. MLMN~\cite{2021Learning} extracted all the relevant law articles for each fact in a law case.
   \end{itemize}

\textbf{Charge Prediction Task}.
As seen in Table \ref{table00}, there are also already many charge prediction tasks and their available datasets~\cite{2018Legal,2020Iteratively,2018few-shot,qiu2020easyaug} in the juridical field. Furthermore, three cases are summarized:
\begin{itemize}
    \item Case-charge correspondence. Two cases are conducted: (1) One-vs-one. To decrease the complexity of judgment prediction, TOPJUDGE-CJO~\cite{2018Legal}, TOPJUDGE-PKU~\cite{2018Legal},  TOPJUDGE-CAIL~\cite{2018Legal}, Criminal-S~\cite{2018few-shot}, Criminal-M~\cite{2018few-shot}, and Criminal-L~\cite{2018few-shot} removed the cases with multiple defendants and multiple charges. (2) One-vs-many. 3.56\%~of cases with more than on charges contained in FLA~\cite{2017Learning} only consider the cases with one defendant, while 2.97\%~of defendants charged differently are included in MAMD~\cite{RN200}.
     \item Common-vs-Few shot. To examine the performance of methods on few-shot charges, all of the Criminal-S~\cite{2018few-shot}, Criminal-M~\cite{2018few-shot}, and Criminal-L~\cite{2018few-shot} keep 149 separate charges with frequencies greater than 10. While FLA~\cite{2017Learning} chooses to consider 50 different charges with frequencies greater than 80, the frequencies of charges greater than 100 are considered in TOPJUDGE-CJO~\cite{2018Legal}, TOPJUDGE-PKU~\cite{2018Legal}, and TOPJUDGE-CAIL~\cite{2018Legal}. 
    \item Logical knowledge. Three cases are conducted: (1) Dependencies among LJP tasks. TOPJUDGE-CJO~\cite{2018Legal}, TOPJUDGE-PKU~\cite{2018Legal}, and TOPJUDGE-CAIL~\cite{2018Legal} are three datasets to verify the efficiency of the model based on the topological dependencies among law articles, charges, and prison term prediction. Similarly, the influence of the prediction of law articles on charge prediction is validated in~\cite{2017Learning}. (2) Domain knowledge. The pre-defined attributes of the disciminative charges~\cite{2018few-shot} have been utilized in the charge prediction. Coincidentally, the answering of predefined element questions to assist the interpretable charge prediction has been visualized~\cite{2020Iteratively}. (3) Fact-side representation. The charge prediction~\cite{2018Interpretable} is interpreted based on the extraction of Phrase-level facts. Furthermore, the relationship between the sentence-level fact and its corresponding charges has been modeled to the document-level charge prediction~\cite{paul-etal-2020-automatic}.
   \end{itemize}

\textbf{Prison Term Prediction Task}.
Given the present numbers of datasets, as shown in Fig.~\ref{fig:distribution}, the inadequacy of Prison term prediction datasets should be modified. CPTP~\cite{2019Charge}, TOPJUDGE-CJO~\cite{2018Legal}, TOPJUDGE-PKU~\cite{2018Legal} and TOPJUDGE-CAIL~\cite{2018Legal} are datasets built to predict the prison terms according to the relevant charges.    

\textbf{Court View generation Task}.
Besides the law articles, the Court View is another explanation written by judges to interpret the judgment of a case in a country with the civil law system. Due to the difficulty of Court View generation task arising from complex logical reasoning about facts and relevant law articles, there are not too many papers on this topic, as shown in Fig. \ref{fig:distribution}. Two cases are concluded: (1) Dependencies of charge-label and Court View. Court-View-Gen~\cite{Hai2018Interpretable} has utilized charge-label to generate the charge-discriminative Court View for criminal cases. (2) Fact-side representation. AC-NLG~\cite{wu-etal-2020-de} has developed to generate Court Views for civil cases based on a pleas-aware side facts representation.

\textbf{Plea Judgment Task}.
Plea judgment task for court cases generally uses the fact description to predict its judgments on plaintiffs' pleas. Three cases are concluded: (1) Judge-summarized facts narrative. For example, Auto-Judge~\cite{2018Automatic} and ILDC~\cite{2021JUSTICE2021} have been present to the judges' final pro or con decision based on a text summary of the absolute neutral facts of each case. (2) Facts from court debate. LJP-MSJudge~\cite{RN202} predicts the verdicts of the  plaintiffs' pleas based on the facts from real court debate. (3) Case material except for the plea judgments. To predict interpretable plea judgments based on case material masked with the plea judgments has been present in~\cite{malik2021ildc,strickson2020legal,sulea2017predicting}.

\section{Datasets}\label{Datasets}
\begin{table*}
\centering
\caption{The statistics of the existing LJP tasks. Here FSCS: Federal Supreme Court of Switzerland, FSC: French Supreme Court, UKC: UK’s highest court, SCOTUS: Supreme Court of the United States, SCI: Supreme Court of Indian, TSC: Thai Supreme Court, ECHR: European Court of Human Rights, SPC: Supreme People’s Court of China.}
\label{table2}
\setlength{\tabcolsep}{4pt}
\begin{tabular}{|c|c|cc|c|c|c|c|}
\hline
\multirow{2}{*}{Dataset}                                                  & \multirow{2}{*}{Source}                               & \multicolumn{2}{c|}{Category}                                                    & \multirow{2}{*}{\#Cases} & \multirow{2}{*}{\#Articles} & \multirow{2}{*}{\#Charges} & \multirow{2}{*}{\#Prison terms} \\ \cline{3-4}
                                                                          &                                                       & \multicolumn{1}{c|}{Law Domain}                    & Legal System                &                          &                             &                            &                                 \\ \hline
BStricks\_LDC~\cite{strickson2020legal}             & UKC                                                   & \multicolumn{1}{c|}{\multirow{3}{*}{Generic}}      & \multirow{4}{*}{Common-Law} & 4,959                    & \multirow{4}{*}{-}          & \multirow{3}{*}{-}         & \multirow{6}{*}{-}              \\ \cline{1-2} \cline{5-5}
JUSTICE~\cite{2021JUSTICE2021}                      & \begin{tabular}[c]{@{}c@{}}SCUS\\ (Oyez)\end{tabular} & \multicolumn{1}{c|}{}                              &                             & 2,384                    &                             &                            &                                 \\ \cline{1-2} \cline{5-5}
ILDC~\cite{malik2021ildc}                           & \multirow{2}{*}{SCI}                                  & \multicolumn{1}{c|}{}                              &                             & 34,816                   &                             &                            &                                 \\ \cline{1-1} \cline{3-3} \cline{5-5} \cline{7-7}
ACI~\cite{paul-etal-2020-automatic}                 &                                                       & \multicolumn{1}{c|}{\multirow{2}{*}{Criminal}}     &                             & 4,338                    &                             & 20                         &                                 \\ \cline{1-2} \cline{4-7}
TSCC~\cite{kowsrihawat2018predicting}               & TSC                                                   & \multicolumn{1}{c|}{}                              & \multirow{29}{*}{Civil-Law} & 1,207                    & 122                         & \multirow{3}{*}{-}         &                                 \\ \cline{1-3} \cline{5-6}
DPAM~\cite{RN224}                                   & \multirow{23}{*}{SPC}                                 & \multicolumn{1}{c|}{\multirow{18}{*}{Criminal}}    &                             & 17,160                   & 70                          &                            &                                 \\ \cline{1-1} \cline{5-6} \cline{8-8} 
MLMN~\cite{2021Learning}                            &                                                       & \multicolumn{1}{c|}{}                              &                             & 1,189                    & 86                          &                            & 5                               \\ \cline{1-1} \cline{5-8} 
CAIL2018~\cite{xiao2018cail2018}                    &                                                       & \multicolumn{1}{c|}{}                              &                             & 2,676,075                & 183                         & 202                        & 202                             \\ \cline{1-1} \cline{5-8} 
TOPJUDGE-CJO~\cite{2018Legal}                       &                                                       & \multicolumn{1}{c|}{}                              &                             & 1,007,744                & 98                          & 99                         & 11                              \\ \cline{1-1} \cline{5-8} 
TOPJUDGE-PKU~\cite{2018Legal}                       &                                                       & \multicolumn{1}{c|}{}                              &                             & 175,744                  & 68                          & 64                         & 11                              \\ \cline{1-1} \cline{5-8} 
TOPJUDGE-CAIL~\cite{2018Legal}                      &                                                       & \multicolumn{1}{c|}{}                              &                             & 113,536                  & 105                         & 122                        & 11                              \\ \cline{1-1} \cline{5-8} 
FLA~\cite{2017Learning}                             &                                                       & \multicolumn{1}{c|}{}                              &                             & 60,000                   & 321                         & 50                         & \multirow{3}{*}{-}              \\ \cline{1-1} \cline{5-7}
RACP~\cite{2018Interpretable}                       &                                                       & \multicolumn{1}{c|}{}                              &                             & 100,000                  & \multirow{7}{*}{-}          & 51                         &                                 \\ \cline{1-1} \cline{5-5} \cline{7-7}
MAMD~\cite{RN200}                                   &                                                       & \multicolumn{1}{c|}{}                              &                             & 164,997                  &                             & -                          &                                 \\ \cline{1-1} \cline{5-5} \cline{7-8} 
CPTP~\cite{2019Charge}                              &                                                       & \multicolumn{1}{c|}{}                              &                             & 238,749                  &                             & 157                        & 226                             \\ \cline{1-1} \cline{5-5} \cline{7-8} 
Court-View-Gen~\cite{Hai2018Interpretable}          &                                                       & \multicolumn{1}{c|}{}                              &                             & 171,981                  &                             & 51                         & \multirow{7}{*}{-}              \\ \cline{1-1} \cline{5-5} \cline{7-7}
Criminal-S~\cite{2018few-shot}                      &                                                       & \multicolumn{1}{c|}{}                              &                             & 77,046                   &                             & 149                        &                                 \\ \cline{1-1} \cline{5-5} \cline{7-7}
Criminal-M~\cite{2018few-shot}                      &                                                       & \multicolumn{1}{c|}{}                              &                             & 191,960                  &                             & 149                        &                                 \\ \cline{1-1} \cline{5-5} \cline{7-7}
Criminal-L~\cite{2018few-shot}                      &                                                       & \multicolumn{1}{c|}{}                              &                             & 383,697                  &                             & 149                        &                                 \\ \cline{1-1} \cline{5-7}
QAjudge-CJO~\cite{2020Iteratively}                  &                                                       & \multicolumn{1}{c|}{}                              &                             & 15,120                   & 19                          & 20                         &                                 \\ \cline{1-1} \cline{5-7}
QAjudge-PKU~\cite{2020Iteratively}                  &                                                       & \multicolumn{1}{c|}{}                              &                             & 14,000                   & 19                          & 20                         &                                 \\ \cline{1-1} \cline{5-7}
QAjudge-CAIL~\cite{2020Iteratively}                 &                                                       & \multicolumn{1}{c|}{}                              &                             & 13,423                   & 19                          & 20                         &                                 \\ \cline{1-1} \cline{5-8} 
\multirow{2}{*}{CAIL-Long~\cite{xiao2021lawformer}} &                                                       & \multicolumn{1}{c|}{}                              &                             & 115,849                  & 244                         & 201                        & 240                             \\ \cline{3-3} \cline{5-8} 
                                                                          &                                                       & \multicolumn{1}{c|}{\multirow{5}{*}{Civil}}        &                             & 113,656                  & 330                         & 257                        & \multirow{10}{*}{-}             \\ \cline{1-1} \cline{5-7}
DPAM~\cite{RN224}                                   &                                                       & \multicolumn{1}{c|}{}                              &                             & 4,033                    & 30                          & \multirow{9}{*}{-}         &                                 \\ \cline{1-1} \cline{5-6}
LJP-MSJudge~\cite{RN202}                            &                                                       & \multicolumn{1}{c|}{}                              &                             & 70,482                   & \multirow{2}{*}{-}          &                            &                                 \\ \cline{1-1} \cline{5-5}
AC-NLG~\cite{wu-etal-2020-de}                       &                                                       & \multicolumn{1}{c|}{}                              &                             & 66,904                   &                             &                            &                                 \\ \cline{1-1} \cline{5-6}
Auto-Judge~\cite{2018Automatic}                     &                                                       & \multicolumn{1}{c|}{}                              &                             & 100,000                  & 62                          &                            &                                 \\ \cline{1-3} \cline{5-6}
echr~\cite{RN220}                                   & \multirow{3}{*}{ECtHR}                                & \multicolumn{1}{c|}{\multirow{3}{*}{Human rights}} &                             & 584                      & 3                           &                            &                                 \\ \cline{1-1} \cline{5-6}
ECHR-CASES~\cite{2019Neural}                        &                                                       & \multicolumn{1}{c|}{}                              &                             & 11,478                   & 66                          &                            &                                 \\ \cline{1-1} \cline{5-6}
ECtHR\_crystal\_ball~\cite{2018Medvedeva}           &                                                       & \multicolumn{1}{c|}{}                              &                             & 11,532                   & 14                          &                            &                                 \\ \cline{1-3} \cline{5-6}
SwissJudgmentPrediction~\cite{niklaus2021swiss}     & FSCS                                                  & \multicolumn{1}{c|}{\multirow{2}{*}{Generic}}      &                             & 85,268                   & \multirow{2}{*}{-}          &                            &                                 \\ \cline{1-2} \cline{5-5}
Sulea et al.~\cite{sulea2017predicting}             & FSC                                                   & \multicolumn{1}{c|}{}                              &                             & 126,865                  &                             &                            &                                 \\ \hline
\end{tabular}
\end{table*}

A significant challenge for LJP is the availability and potentiality of the datasets. In constructing datasets, there have been at least two lines of work, as seen in Table \ref{table00}. The first one employs a machine to extract the necessary metadata, such as using Regular Expression to identify and mask judgment results (~\cite{niklaus2021swiss,sulea2017predicting,strickson2020legal,malik2021ildc}) or to extract the input-output data samples(~\cite{RN220,2019Neural,2018Medvedeva,RN224,xiao2018cail2018,2018Legal,xiao2021lawformer,2017Learning,RN200,Hai2018Interpretable,2019Charge,2018Automatic}) from judgment documents of given court  and applying the deep neural model to extract facts from Court judgment documents as~\cite{paul-etal-2020-automatic} for evaluating the LJP model. The second line of work recruits legal experts to annotate rationale sentences~\cite{malik2021ildc,2018Interpretable} as gold sentences or to annotate legal domain knowledge (which are not included in the judgment documents, such as ~\cite{2021Learning,2018few-shot,2020Iteratively}).

Specially, we categorize public LJP datasets into two categories: task-specific datasets and multi-task datasets. Multi-task datasets are composed of multiple subtasks of LJP, while single-task datasets consist of a single subtask of LJP.

\subsection{Single-Task Datasets}
The following list outlines the different types of task-specific datasets released publicly. They are categorized based on their task's outcomes.
\subsubsection{Article Prediction  Datasets}
HUDOC European Court of Human Rights (ECHR)\footnote{https://hudoc.echr.coe.int/} and China Judgments Online (CJO)\footnote{https://wenshu.court.gov.cn/} are two publicly available case databases, which include judgment, verdict, conciliation statement, decision letter, notice, et., published by the ECHR and the Supreme People’s Court of China, respectively. 

There are several datasets for articles related to tasks scraped from publicly available resources to unveil patterns driving judicial decisions, but they tend to be less related to the civil and administrative cases (see Table \ref{table2}).

Dataset echr~\cite{RN220} is the first public English legal judgment prediction dataset, containing a total of 584 cases from the ECHR and  articles 3, 6, and 8 from the European Convention of Human Rights, to judge whether the covering case has violated an
article of the convention of human rights. Similarly, Thai Supreme Court Cases (TSCC) ~\cite{kowsrihawat2018predicting}, containing 1,207 criminal judgment records and 122 Law records from the Supreme Court of Thailand, is constructed to predict which specific Law records have been violated based on sequence models. Then, a substantially more extensive data set, ECHR-CASES~\cite{2019Neural}, has been constructed based on 11,478 cases tried by the ECHR and 66 articles from the European Convention of Human Rights, not only to judge whether there has been a violation of an
article of the convention of human rights but also to determine the name of the violated articles.
To verify the improvement on articles recommendation accuracy through establishing the fine-grained fact-article correspondences, a fact-article correspondence dataset, MLMN~\cite{2021Learning}, has been created based on 1,189 Crime judgment documents in CJO and 86 criminal law articles for fact-article correspondence annotation.
\subsubsection{Charge Prediction Datasets}
Case Information
Disclosure (CID)\footnote{http://www.ajxxgk.jcy.gov.cn/html/index.html}, CJO are two publicly available case databases published by the Supreme People’s prosecution of China and the Supreme People’s Court
of China, respectively. 

There have been several datasets for charge prediction tasks scraped from publicly available resources, but they tend to focus on Chinese mainly and have less considered in English (see Table~\ref{table2}).

From CJO, FLA~\cite{2017Learning} collects 60,000 cases in total, 50 charges,  383 words per fact description averagely, 3.81 articles per case, and 3.56\% cases with more than one charge, 321 distinct articles, retains the cases with one defendant and treats the charges that appeared more than 80 times as positive data, vice versa as negative ones, to improve the charge prediction followed the prediction of the relevant law articles. CAIL2018~\cite{2018CAIL2018} is the first large-scale Chinese legal dataset to predict relevant law articles, charges, and prison terms, respectively. It includes 2,676,075 criminal cases published by the Supreme People’s Court
of China, 183 criminal law articles, 202 charges, and prison terms. And this dataset only retains the cases with a single defendant and treats those charges and law articles whose frequency is larger than 30 as positive data. 
MAMD~\cite{RN200} used to predict multi-defendant charges has been built from the published legal documents in Case Information
Disclosure (CIS), containing a total of 164,997 cases in which fact description, defendants’ names, and charges contained. Cases involving multiple defendants account for about 30$\%$, and in cases involving multiple defendants,
the ratio of that all defendants are charged the same is about 90$\%$.
RACP~\cite{2018Interpretable}, has been constructed from CJO, containing 100,000 documents in which rationale sentences were annotated based on the extracted fact description and charge labels. Similarly, ACI~\cite{paul-etal-2020-automatic} collected from the Supreme Court of India cases consists of 4,338 judgment documents with document-level charge label information facts were extracted by legal experts and automated method for 70 and 4,268 documents, and legal experts annotate sentence-level charges for 120 documents. Moreover, three datasets~\cite{2018few-shot} have been published with selected case’s fact part and extracted charges of judgment documents from CJO denoted as Criminal-S(small), Criminal-M(medium), Criminal-L(large). They removed the cases which have more than one charge in a verdict. 
\subsubsection{Prison Term Prediction  Datasets}
 CJO is the publicly available case database published by the Supreme People’s Court
of China, including judgment, verdict, conciliation statement, decision letter, notice, and so on. 

Several datasets on prison term prediction scraped from publicly available resources, mainly focusing on Chinese and rarely English (see Table \ref{table2}).

\begin{table*}
\centering
\caption{The statistics of the pre-training datasets.}
\label{table7}
\setlength{\tabcolsep}{7pt}
\resizebox{\textwidth}{1.8in}{
\begin{tabular}{|c|c|c|c|c|c|c|}
\hline
Dataset                                                                   & Language                  & Model                                                                                                                                & Source                                                                               & \#Documents              & \#Size(MB)           & Corpus              \\ \hline
\multirow{6}{*}{Chalkidis et al.~\cite{2020LEGAL}}        & \multirow{19}{*}{English} & \multirow{6}{*}{\href{https:
//huggingface.co/nlpaueb}{LEGAL-BERT~\cite{2020LEGAL}}}                                                                                                          & EU legislation                                                                       & 61,826                   & 1,900                  & \multirow{6}{*}{-}    \\ \cline{4-6}
                                                                          &                           &                                                                                                                                      & European Court of Justice (ECJ) cases                                               & 19,867                   & 600                  &                       \\ \cline{4-6}
                                                                          &                           &                                                                                                                                      & UK legislation                                                                       & 19,867                   & 1,400                  &                       \\ \cline{4-6}
                                                                          &                           &                                                                                                                                      & ECHR cases                                                                           & 12,554                   & 500                  &                       \\ \cline{4-6}
                                                                          &                           &                                                                                                                                      & US court cases                                                                       & 164,141                  & 3,200                  &                       \\ \cline{4-6}
                                                                          &                           &                                                                                                                                      & US contracts                                                                         & 76,366                   & 3,900                  &                       \\ \cline{1-1} \cline{3-7} 
\multirow{7}{*}{LexGLUE~\cite{chalkidis2021lexglue}}      &                           & \multirow{7}{*}{\begin{tabular}[c]{@{}c@{}}\href{https://github.com/
google-research/bert}{BERT}~\cite{devlin-etal-2019-bert}\\ \href{https://github.com/pytorch/fairseq}{RoBERTa}~\cite{liu2019roberta}\\ \href{https://github.com/microsoft/DeBERTa}{DeBERTa}~\cite{he2020deberta}\\ \href{https://github.com/allenai/longformer}{Longformer}~\cite{beltagy2020longformer}\\ \href{http://goo.gle/bigbird-transformer}{BigBird}~\cite{zaheer2020big}\\ \href{https:
//huggingface.co/nlpaueb}{LEGAL-BERT~\cite{2020LEGAL}}\\ \href{https://github.com/reglab/
casehold}{CaseLaw-BERT}~\cite{zhengguha2021}\end{tabular}} & ECHR~\cite{2019Neural}                                                                                 & 1,1000                   & 116                    & \multirow{7}{*}{\href{https://huggingface.co/datasets/lex_glue}{Link}} \\ \cline{4-6}
                                                                          &                           &                                                                                                                                      & ECHR~\cite{chalkidis2021paragraph}                                                                                 & 1,1000                   & 116                    &                       \\ \cline{4-6}
                                                                          &                           &                                                                                                                                      & US Law~\cite{spaeth2020supreme}                                                                               & 7,800                    & 328                    &                       \\ \cline{4-6}
                                                                          &                           &                                                                                                                                      & EU Law~\cite{chalkidis2021multieurlex}                                                                               & 65,000                   & 492                    &                       \\ \cline{4-6}
                                                                          &                           &                                                                                                                                      & Contracts~\cite{tuggener2020ledgar}                                                                            & 80,000                   & 62                    &                       \\ \cline{4-6}
                                                                          &                           &                                                                                                                                      & Contracts~\cite{lippi2019claudette}                                                                            & 9,414                    & 3                    &                       \\ \cline{4-6}
                                                                          &                           &                                                                                                                                      & Harvard Law case~\cite{zhengguha2021}                                                                     & 52,800                   & 86                    &                       \\ \cline{1-1} \cline{3-7} 
CaseHOLD~\cite{zhengguha2021}                             &                           & \href{https://github.com/reglab/
casehold}{CaseLaw-BERT}~\cite{zhengguha2021}                                                                                                                         & Harvard Law case~\cite{zhengguha2021}                                                                     & 3,446,187                & 37,000                   & \href{https://github.com/reglab/
casehold}{Link}                  \\ \cline{1-1} \cline{3-7} 
\multirow{5}{*}{CourtListener~\cite{bambroo2021legaldb}} &                           & \multirow{5}{*}{LegalDB~\cite{bambroo2021legaldb}}                                                                                                             & US Board Of Tax Appeal                                                               & 11,059                   & \multirow{5}{*}{8,000}   & \multirow{5}{*}{\href{https://www.courtlistener.com/api/bulk-info/}{Link}} \\ \cline{4-5}
                                                                          &                           &                                                                                                                                      & US Court Of Federal Claims                                                           & 13,410                   &                      &                       \\ \cline{4-5}
                                                                          &                           &                                                                                                                                      & Court Of Customs And Patents Appeal                                                  & 2,388                    &                      &                       \\ \cline{4-5}
                                                                          &                           &                                                                                                                                      & Supreme Court Of The United States                                                   & 31,470                   &                      &                       \\ \cline{4-5}
                                                                          &                           &                                                                                                                                      & \begin{tabular}[c]{@{}c@{}}Court Of Appeals (First–Eleventh Circuit)\end{tabular} & 269,622                  &                      &                       \\ \hline
\multirow{2}{*}{Xiao et al.~\cite{xiao2021lawformer}}         & \multirow{2}{*}{Chinese}  & \multirow{2}{*}{\href{https://github.
com/thunlp/LegalPLMs}{Lawformer}}                                                                                                           & SPC Criminal cases                                                                   & 5,428,717                & 17,000                   & \multirow{2}{*}{-}    \\ \cline{4-6}
                                                                          &                           &                                                                                                                                      & SPC Civil cases                                                                      & 17,387,874               & 67,000                   &                       \\ \hline
\multirow{2}{*}{Douka et al.~\cite{2021Douka}}           & \multirow{2}{*}{French}   & \multirow{2}{*}{\href{http://master2-bigdata.polytechnique.fr/resources\#juribert}{JuriBERT}}                                        & French Court cases   & \multirow{2}{*}{123,361} & \multirow{2}{*}{6,300} & \multirow{2}{*}{-}    \\ \cline{4-4}
     &                      &                          & Legifrance             &              &      &           \\ \hline
\end{tabular}}
\end{table*}
To verify the influence of the charge-based prison term prediction (CPTP) on the prediction accuracy of the full prison term for each defendant, CPTP~\cite{2019Charge} has been constructed with 238,749 criminal cases [1,240] (in months) prison terms, and 157 types of charges. The fact-article correspondence dataset, MLMN, constructed in~\cite{2021Learning} can also boost the downstream task of legal decision prediction, where all judgment results are divided into five classes: exempt from
criminal punishment, criminal detention, fixed-term imprisonment
of not more than one year, 1 - 3 years and not less than three years.
\subsubsection{Plea Judgment prediction  Datasets}
CJO, entscheidsuche.ch, IndianKanoon, Court de Cassation, and Oyez are the publicly available source case databases published by the Supreme People’s Court of China (SPC), the Federal Supreme Court of Switzerland (FSCS), the Supreme Court of India (SCI), the French Supreme Court and the Supreme Court of the United States (SCOTUS), respectively. 

Several datasets for pleas-related tasks are scraped from publicly available resources, but these large-scale datasets mainly focus on SPC and FSC cases (see Table \ref{table2}).

For example, to predict the final judgment results based on the semantic interactions among facts, pleas, and laws, Auto-Judge~\cite{2018Automatic} consists of 100,000 divorce cases, 185,723 pleas and verdicts, and 62 law articles. To employ the input of the case from real courtrooms rather than the judge-summarized case narrative for the judgment prediction, LJP-MSJudge~\cite{RN202}, has been released with 70,482 Private Lending cases collected from CJO, 133,209 claims and verdicts, 4.1 million debates, and ten fact labels. Furthermore, Sulea et al.~\cite{sulea2017predicting} proposed a dataset comprising 126,865 unique court rulings firstly used to predict court rulings in the French Supreme Court cases. Indian Legal Documents Corpus (ILDC~\cite{malik2021ildc}) with 35k cases is introduced for court judgment prediction and explanation. Moreover, a labeled data set of 4,959 UK court cases~\cite{strickson2020legal} is also created for legal judgment prediction. A total of 2,384 SCOTUS cases from Oyez based on a series of manually balance procedures are also evaluated to predict the judicial judgment. In addition, SwissJudgmentPrediction~\cite{niklaus2021swiss} is a diachronic multilingual dataset of 85K cases from the Federal Supreme Court of Switzerland (FSCS), where 50k cases in German, 31k in French, and 4K in Italian.

\subsubsection{Court view Generation  Datasets}
CJO is the publicly available case database published by the Supreme People’s Court of China, including judgment, verdict, conciliation statement, decision letter, notice, and so on. 

Several datasets for the generation tasks of rationales in the court view scraped from publicly available resources, but they tend to mainly focus on Chinese rather than English (see Table \ref{table00}).

Court-View-Gen~\cite{Hai2018Interpretable} has been constructed from the published legal documents in CJO, from which they collected 171,981 cases with one defendant and one charge, a total of 51 charge labels, to explore the generation of court views based on charge labels.  
\subsection{Multi-Task Datasets}
Multi-task datasets ( see Table \ref{table00} ) are the datasets that contain detailed and rich subtasks of judgment predictions to help the researchers make improvements on legal judgment prediction.

For example, datasets QAjudge-CJO, QAjudge-PKU, and QAjudge-CAIL~\cite{2020Iteratively} have been gathered from China Judgments Online, Peking University Law Online, and Chinese AI and Law Challenge, respectively. These datasets consist of fact descriptions, applicable law articles, charges, and the term of penalty for each case. These three datasets filtered out cases with multiple defendants, multiple charges, infrequent charges, and articles less than 100 times. 

Furthermore, the large-scale publicly accessible high-quality Chinese legal judgment prediction dataset, CAIL2018~\cite{2018CAIL2018} has been released with 2,676,075 criminal cases, 183 criminal law articles, 202 charges, and prison terms. This dataset only retained the cases with a single defendant, those charges, and law articles whose frequency is larger than 30. 

Recently, a much larger scale dataset, CAIL-Long ~\cite{xiao2021lawformer}, was constructed to predict the judgment results. This dataset consists of 1,129,053 criminal
cases and 1,099,605 civil cases. Specifically, each criminal
case is annotated with the charges, the relevant
laws, and the penalty term. Each civil case is
annotated with the causes of actions and the relevant laws. 

In addition, from the perspective of court view generation, the joint generation of the judgment and the rationales as court view can improve the interpretability for the judgment of a case, where the judgment is a response to the plaintiff's claims. For example, AC-NLG~\cite{2020De} built 66,904 cases based on civil legal judgment documents from CJO, in which each case was categorized into plaintiff’s claim, facts description, and court’s view was further annotated the judgments and rationales.
\subsection{Datasets for Pre-trained Language Model}
As is well known, there have been many open access repositories to construct unlabelled pretraining corpora, as shown in Table~\ref{table7}. For example, CJO and HUDOC ECHR are publicly available case databases published by the Supreme People’s Court of China and ECHR. Furthermore, CourtListener\footnote{https://www.courtlistener.com/api/bulk-info/} and Légifrance websites consist of US court cases and raw French legal text. In addition, access to the Case Law Access Project is granted upon request. 

Inspired by the success of pre-trained language models (PLMs) in the generic domain, there have been several datasets scraped from publicly available legal resources to investigate the adaptation of PLMs to legal tasks as follows.

For example, 12GB pre-trained unlabelled corpora of diverse English legal text~\cite{2020LEGAL} from several fields (e.g., legislation, court cases, contracts) has been scraped from publicly available resources to pre-train the different variations of BERT in the legal area. Then, to push towards a generic benchmark dataset for multiple legal NLP tasks in English, the Legal General Language Understanding Evaluation (LexGLUE~\cite{chalkidis2021lexglue}) is selected using criteria primarily from SuperGLUE. Moreover, 8GB of long legal documents from the US are also used as a legal domain pre-trained corpora to validate the efficiency of the LegalDB model~\cite{bambroo2021legaldb}. A large-scale pretraining dataset, Case Holdings On Legal Decisions (CaseHOLD), size 37GB with 3,446,187 legal decisions from Harvard Law case corpus, is constructed to explore the influence of difficulty and domain specificity on domain
pretraining gains. 

While PLMs have proven very useful when adapted to the legal domain, the main effort focuses on the English language. Therefore,  to utilizing PLMs to address legal judgment prediction tasks based on capturing the long-distance dependency on Chinese legal case documents up to 512 tokens, 84GB pre-trained unlabelled corpora~\cite{xiao2021lawformer} of Chinese legal cases scraped from CJO, and a large scale judgment prediction dataset, CAIL-Long has also been constructed(see Table \ref{table00}). In addition,  a
collection of a size of 6.3 GB raw French legal text is collected to explore the performance of the adaptation of domain-specific BERT models in the French language. 

However, all these datasets for legal PLMs tend to focus only on a single language corpus and have less considered a multi-language corpus (see Table \ref{table7}).

\section{Evaluation Metrics}\label{Evaluation Metrics}
From Fig. \ref{Eva}, we can obtain at least four categories for evaluating the outputs from LJP tasks, (e.g. text classification metrics for evaluating law articles, charges, plea judgments and Element answers; text classification with text regression metrics for evaluating prison terms according to its value distribution; text classification with text generation metrics for evaluating Fact snippets based on its generation methods; text generation metrics for evaluating the generated Court View.) 
\subsection{Evaluation of Text Classification}
\begin{figure}
 \centering
  \includegraphics[width=3.2in]{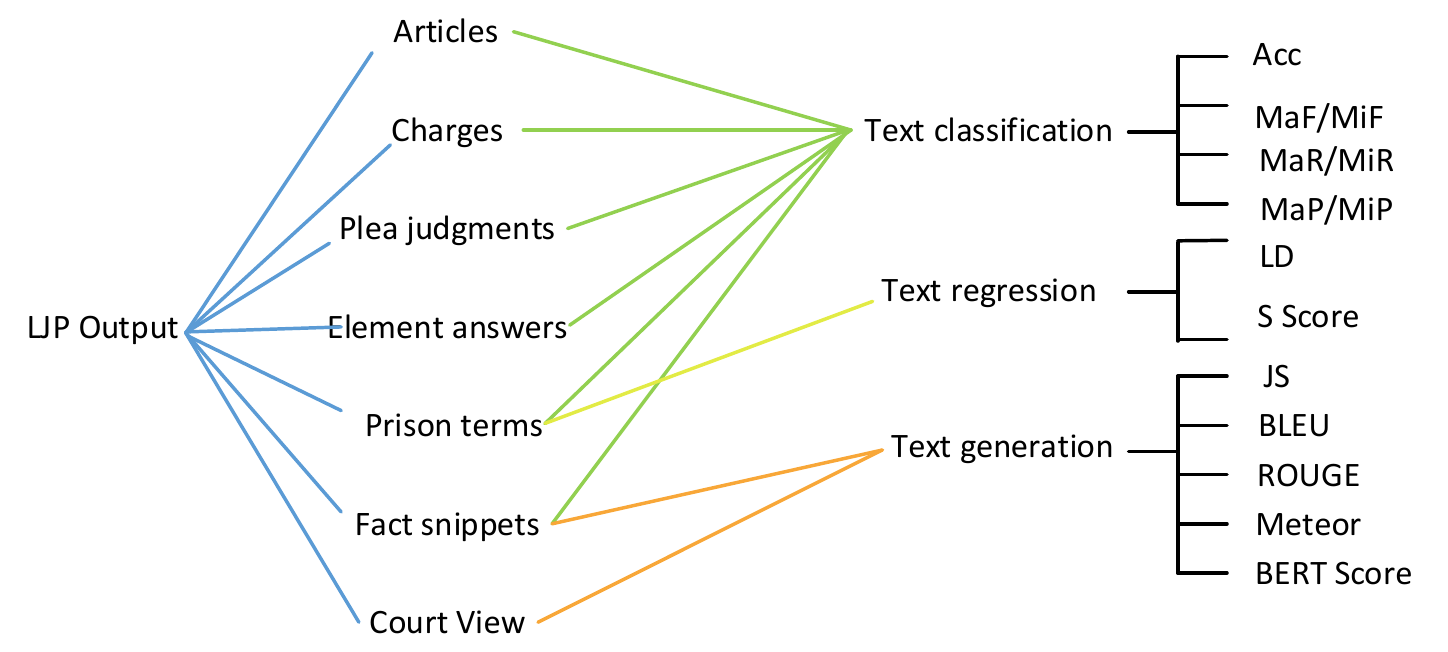}
  \caption{Evaluation metrics for different outputs from LJP tasks.  }\label{Eva}
\end{figure}

\begin{table*}
\centering
\caption{Evaluation metrics used in papers which construct datasets for evaluating the performance of the existing LJP models. Here, EM: exact match rate, Acc@p: error-tolerant accuracy, where p is the maximum acceptable error rate.}
\label{paperlist}
\setlength{\tabcolsep}{4pt}
\begin{tabular}{|c|c|cccc|}
\hline
Dataset                                                                                 & Output                          & \multicolumn{4}{c|}{Evaluation metrics}                                                                                                                      \\ \hline
ECtHR\_crystal\_ball~\cite{2018Medvedeva}                                   & \multirow{2}{*}{Articles}       & \multicolumn{4}{c|}{\multirow{3}{*}{Acc}}                                                                                                                    \\ \cline{1-1}
echr~\cite{RN220}                                            &                                 & \multicolumn{4}{c|}{}                                                                                                                                        \\ \cline{1-2}
MAMD~\cite{RN200}                                                & Charges                         & \multicolumn{4}{c|}{}                                                                                                                                        \\ \hline
\multirow{3}{*}{CAIL-Long~\cite{xiao2021lawformer}}                    & Prison terms                    & \multicolumn{4}{c|}{LD}                                                                                                                                      \\ \cline{2-6} 
                                                                                        & Articles                        & \multicolumn{4}{c|}{\multirow{4}{*}{MaF/MiF}}                                                                                                                \\ \cline{2-2}
                                                                                        & Charges                         & \multicolumn{4}{c|}{}                                                                                                                                        \\ \cline{1-2}
SwissJudgmentPrediction~\cite{niklaus2021swiss}                                 & Plea judgments                  & \multicolumn{4}{c|}{}                                                                                                                                        \\ \cline{1-2}
TSCC~\cite{kowsrihawat2018predicting}                                  & Articles                        & \multicolumn{4}{c|}{}                                                                                                                                        \\ \hline
ACI~\cite{paul-etal-2020-automatic}                             & Charges                         & \multicolumn{2}{c|}{\multirow{3}{*}{MaF}}                                               & \multicolumn{1}{c|}{\multirow{6}{*}{MaP}}  & \multirow{6}{*}{MaR}  \\ \cline{1-2}
ECHR-CASES~\cite{2019Neural}                                     & \multirow{3}{*}{Articles}       & \multicolumn{2}{c|}{}                                                                   & \multicolumn{1}{c|}{}                      &                       \\ \cline{1-1}
MLMN~\cite{2021Learning}                                          &                                 & \multicolumn{2}{c|}{}                                                                   & \multicolumn{1}{c|}{}                      &                       \\ \cline{1-1} \cline{3-4}
DPAM~\cite{RN224}                                                &                                 & \multicolumn{1}{c|}{JS}                    & \multicolumn{1}{c|}{MaF}                   & \multicolumn{1}{c|}{}                      &                       \\ \cline{1-4}
\multirow{2}{*}{RACP~\cite{2018Interpretable}}                 & Charges                         & \multicolumn{2}{c|}{MaF}                                                                & \multicolumn{1}{c|}{}                      &                       \\ \cline{2-4}
                                                                                        & Fact snippets                   & \multicolumn{1}{c|}{\multirow{15}{*}{Acc}} & \multicolumn{1}{c|}{MaF}                   & \multicolumn{1}{c|}{}                      &                       \\ \cline{1-2} \cline{4-6} 
QAjudge-CJO,QAjudge-PKU,QAjudge-CAIL~\cite{2020Iteratively}                 & Elements answers                & \multicolumn{1}{c|}{}                      & \multicolumn{3}{c|}{MaF}                                                                                        \\ \cline{1-2} \cline{4-6} 
JUSTICE~\cite{2021JUSTICE2021}                                          & \multirow{3}{*}{Plea judgments} & \multicolumn{1}{c|}{}                      & \multicolumn{1}{c|}{\multirow{13}{*}{MaF}} & \multicolumn{1}{c|}{\multirow{13}{*}{MaP}} & \multirow{13}{*}{MaR} \\ \cline{1-1}
Sulea et al.~\cite{sulea2017predicting}                                &                                 & \multicolumn{1}{c|}{}                      & \multicolumn{1}{c|}{}                      & \multicolumn{1}{c|}{}                      &                       \\ \cline{1-1}
BStricks\_LDC~\cite{strickson2020legal}                            &                                 & \multicolumn{1}{c|}{}                      & \multicolumn{1}{c|}{}                      & \multicolumn{1}{c|}{}                      &                       \\ \cline{1-2}
\multirow{3}{*}{CAIL2018~\cite{xiao2018cail2018}}                      & Articles                        & \multicolumn{1}{c|}{}                      & \multicolumn{1}{c|}{}                      & \multicolumn{1}{c|}{}                      &                       \\ \cline{2-2}
                                                                                        & Charges                         & \multicolumn{1}{c|}{}                      & \multicolumn{1}{c|}{}                      & \multicolumn{1}{c|}{}                      &                       \\ \cline{2-2}
                                                                                        & Prison terms                    & \multicolumn{1}{c|}{}                      & \multicolumn{1}{c|}{}                      & \multicolumn{1}{c|}{}                      &                       \\ \cline{1-2}
\multirow{3}{*}{\begin{tabular}[c]{@{}c@{}}TOPJUDGE-CJO~\cite{2018Legal}\\TOPJUDGE-PKU~\cite{2018Legal}\\TOPJUDGE-CAIL~\cite{2018Legal}\\ \end{tabular}}      & Articles                        & \multicolumn{1}{c|}{}                      & \multicolumn{1}{c|}{}                      & \multicolumn{1}{c|}{}                      &                       \\ \cline{2-2}
                                                                                        & Charges                         & \multicolumn{1}{c|}{}                      & \multicolumn{1}{c|}{}                      & \multicolumn{1}{c|}{}                      &                       \\ \cline{2-2}
                                                                                        & Prison terms                    & \multicolumn{1}{c|}{}                      & \multicolumn{1}{c|}{}                      & \multicolumn{1}{c|}{}                      &                       \\ \cline{1-2}
Criminal-S,Criminal-M,Criminal-L~\cite{2018few-shot}                   & Charges                         & \multicolumn{1}{c|}{}                      & \multicolumn{1}{c|}{}                      & \multicolumn{1}{c|}{}                      &                       \\ \cline{1-2}
\multirow{2}{*}{\begin{tabular}[c]{@{}c@{}}QAjudge-CJO, QAjudge-PKU,\\QAjudge-CAIL~\cite{2020Iteratively}\\ \end{tabular}} & Articles                        & \multicolumn{1}{c|}{}                      & \multicolumn{1}{c|}{}                      & \multicolumn{1}{c|}{}                      &                       \\ \cline{2-2}
                                                                                        & Charges                         & \multicolumn{1}{c|}{}                      & \multicolumn{1}{c|}{}                      & \multicolumn{1}{c|}{}                      &                       \\ \cline{1-2}
Auto-Judge~\cite{2018Automatic}                                       & Plea judgments                  & \multicolumn{1}{c|}{}                      & \multicolumn{1}{c|}{}                      & \multicolumn{1}{c|}{}                      &                       \\ \hline
FLA~\cite{2017Learning}                                         & Charges                         & \multicolumn{2}{c|}{\multirow{2}{*}{MaF/MiF}}                                           & \multicolumn{1}{c|}{MaP/Mip}               & MaR/MiR               \\ \cline{1-2} \cline{5-6} 
LJP-MSJudge~\cite{RN202}                                                 & \multirow{2}{*}{Plea judgments} & \multicolumn{2}{c|}{}                                                                   & \multicolumn{1}{c|}{\multirow{2}{*}{MaP}}  & \multirow{2}{*}{MaR}  \\ \cline{1-1} \cline{3-4}
\multirow{2}{*}{ILDC~\cite{malik2021ildc}}                             &                                 & \multicolumn{1}{c|}{Acc}                   & \multicolumn{1}{c|}{MaF}                   & \multicolumn{1}{c|}{}                      &                       \\ \cline{2-6} 
                                                                                        & Fact snippets                   & \multicolumn{1}{c|}{JS}                    & \multicolumn{1}{c|}{ROUGE}                 & \multicolumn{1}{c|}{BLEU}                  & Meteor                \\ \hline
Court-View-Gen~\cite{Hai2018Interpretable}                                  & \multirow{2}{*}{Court View}     & \multicolumn{2}{c|}{ROUGE}                                                              & \multicolumn{2}{c|}{\multirow{2}{*}{BLEU}}                         \\ \cline{1-1} \cline{3-4}
AC-NLG~\cite{wu-etal-2020-de}                                        &                                 & \multicolumn{1}{c|}{BERT SCORE}            & \multicolumn{1}{c|}{ROUGE}                 & \multicolumn{2}{c|}{}                                              \\ \hline
CPTP~\cite{2019Charge}                                           & Prison terms                    & \multicolumn{1}{c|}{S}                     & \multicolumn{1}{c|}{EM}                    & \multicolumn{2}{c|}{Acc@p}                                         \\ \hline
\end{tabular}
\end{table*}

To be specific, for the prediction of articles, charges, prison terms, plea judgments, fact snippets and element answers, we can take them as text classification problems, such as in Fig. \ref{Eva}. For a specific text classification task, suppose there are $M$ categories and $N$ law cases. This text classification task aims to predict the category label for the text description of each law case. Let $y_{ij}\in\{0, 1\} (i\in\{1, 2, \cdots, M\}, j\in\{1, 2, \cdots, N\})$ denote as the ground truth label of the category result. Let $\hat{y}_{ij}\in\{0, 1\} (i\in\{1, 2, \cdots, M\}, j\in\{1, 2, \cdots, N\})$ denote as the predict label of the category result. Then, we can obtain the true positive, false positive, false negative, true negative, precision, recall and  metrics for the $i$-th category as follows:
\begin{eqnarray*}
&\hspace{-0.4cm} TP_{i}=\sum_{j=1}^{N}{[y_{ij}=1,\hat{y}_{ij}=1]}, \notag \\
&\hspace{-0.4cm}FP_{i}=\sum_{j=1}^{N}{[y_{ij}=0,\hat{y}_{ij}}=1], \notag \\
&\hspace{-0.4cm}FN_{i}=\sum_{j=1}^{N}{[y_{ij}=1,\hat{y}_{ij}=0]}, \notag \\
&\hspace{-0.4cm}TN_{i}=\sum_{j=1}^{N}{[y_{ij}=0,\hat{y}_{ij}=0]}
, \notag \\
&\hspace{-0.4cm}
P_{i}= \frac{TP_{i}}{TP_{i}+FP_{i}}, \notag
\\
 &\hspace{-0.4cm} R_{i}= \frac{TP_{i}}{TP_{i}+FN_{i}}
\end{eqnarray*}

Then, as in Table \ref{paperlist}, we can obtain the following evaluation metrics to evaluate the performance of LJP text classification.

\begin{itemize}
    \item \textbf{Macro precision/ Macro recall/ Macro F value}. To evaluate the performance in the macro-level through averaging over each category, macro precision $MaP$, macro recall $MaR$ and macro F value $MaF$ are follows:
   \begin{eqnarray}
&\hspace{-0.4cm} MaP=\frac{1}{M}\sum_{i=1}^{M} P_{i} \notag
\\
 &\hspace{-0.4cm} MaR=\frac{1}{M}\sum_{i=1}^{M} R_{i} \notag\\
&\hspace{-0.4cm}  MaF=\frac{1}{M}\sum_{i=1}^{M} \frac{2P_{i}\times R_{i}}{P_{i}+R_{i}}
\end{eqnarray}
  \item \textbf{Micro precision/ Micro recall/ Micro F value/ Acc}. To evaluate the performance in the micro-level through averaging over each law case, micro precision $MiP$, micro recall $MiR$, micro F value $MiF$ and  $Acc$ are follows:
   \begin{eqnarray*}
&\hspace{-0.4cm} MiP=\frac{\sum_{i=1}^{M}TP_{i}}{\sum_{i=1}^{M}TP_{i}+FP_{i}} \notag
\\
 &\hspace{-0.4cm} MiR=\frac{\sum_{i=1}^{M}TP_{i}}{\sum_{i=1}^{M}TP_{i}+FN_{i}} \notag\\
&\hspace{-0.4cm}  MiF= \frac{2P_{micro}\times R_{micro}}{P_{micro}+R_{micro}} \notag\\
&\hspace{-0.4cm}
Acc=\frac{\sum_{i=1}^{M}TP_{i}+TN_{i}}{\sum_{i=1}^{M}TP_{i}+FP_{i}+TN_{i}+FN_{i}}
\end{eqnarray*}
 \end{itemize}   
\subsection{Evaluation of Text Generation}
For the generation of court view, we can take them as text generation problems, such as BLEU-1, BLEU-2, BLEU-N, ROUGE-1, ROUGE-2, ROUGE-L and BERT SCORE~\cite{shen2019select,shen2019improving,zhao2019unsupervised,zhao2018comprehensive,wu-etal-2020-de,su2020diversifying,su2020moviechats,chang2021neural}. The following evaluation metrics to evaluate the performance of LJP text generation are summarized, as shown in Fig. \ref{Eva}.
\begin{itemize}
    \item \textbf{Jaccard similarity}. To evaluate the similarity between two sets, Jaccard similarity is defined as the size of the intersection divided by the size of the union of the two sets, which is follows:
    \begin{equation}
    JS  =\frac{\mid \{Candidates\} \cap \{References\} \mid}{\mid \{Candidates\} \cup \{References\} \mid}
\end{equation}
    where, $\{Candidates\}$ denotes the text set predicted, and $\{Candidates\}$ denotes the text set of the reference.

      \item \textbf{BLEU}. To evaluate the exact form closeness between the candidate sentence and its reference sentences, BLEU is averaged geometrically computed for mutiple modified $n$-gram up to length $N$ precision (e.g. $n=1,2,3,4, N=4$), which is follows:
    \begin{equation}
    BLEU  =BP \cdot \exp\biggl( \sum_{n=1}^{N} w_{n}\log p_{n} \biggr)
\end{equation}
    where $BP$, $p_{n}$, $w_{n}$ are an exponential brevity penalty factor, modified $n$-gram precision and a weighted coefficients of modified $n$-gram precision for the up to $N$-gram combining system, respectively.
        \begin{equation}
    BP =\left\{
\begin{array}{lcl}
1, & & {if \quad c > r,}\\
e^{(1-r/c)}, & & {if \quad c \leq r.}
\end{array}
 \right.
\end{equation}
where $c$ denotes the length of the candidate sentence and $r$ denotes the length of the reference sentences.

            \item \textbf{ROUGE}. To evaluate $n$-gram co-occurrence statistics between the computer-generated text and the referenced text created by humans, a family of ROUGE metrics is defined as a recall-based measure, which is follows:
            \begin{eqnarray*}
 \begin{aligned}
& ROUGE-N \notag
\\
& =\frac{\sum\limits_{S \in \{References\}}\sum\limits_{{gram}_{n} \in S}{Count}_{match}({gram}_{n})}{\sum\limits_{S \in \{References\}}\sum\limits_{{gram}_{n} \in S}{Count}({gram}_{n})}
\end{aligned}   
\end{eqnarray*}

    where the numerator of $ROUGE-N$ denotes the number of $n$-grams co-occurring between a candidate text and a set of reference texts, and the denominator of $ROUGE-N$ denotes the number of $n$-grams in the set of reference texts.
   \begin{eqnarray}
&\hspace{-0.4cm}
R_{lcs}= \frac{LCS(X,Y)}{m} \notag
\\
 &\hspace{-0.4cm} P_{lcs}= \frac{LCS(X,Y)}{n} \notag\\
&\hspace{-0.4cm} \beta=\frac{P_{lcs}}{R_{lcs}} \notag
\\
 &\hspace{-0.4cm} ROUGE-L=\frac{(1+\beta^2)P_{lcs}R_{lcs}}{R_{lcs}+\beta^2P_{lcs}}
\end{eqnarray} 
where $LCS(X,Y)$ is the length of a longest common subsequence of sequence $X$ of length $m$ and sequence $Y$ of length $n$.
    
               \item \textbf{Meteor}. To give an explicit word-to-word matching between the candidate text and its reference texts, Meteor is allowing backing-off from the exact unigram matching to porter stem matching and synonyms, which is follows:
               \begin{equation}
                Meteor=F3\times (1-penalty)   
               \end{equation}
               where $F3$ denotes a harmonic metric by combining the unigram precision ($P$, the ratio of number of unigrams mapped to the total unigrams in the candidate text) and the unigram recall ($R$, the ratio of number of unigrams mapped to the total unigrams in the reference text). And $penalty$ denotes a penal function increases as the number of mapped subsequence chunks while decreasing as the number of unigrams mapped. 
                 \begin{eqnarray*}
               &\hspace{-0.4cm} F3=\frac{10PR}{R+9P} \notag \\
               &\hspace{-0.4cm} penalty=0.5\times { \Big ( \frac{\#chunks}{\#unigrams\_matched}\Big )}^{3}
               \end{eqnarray*}
    
    where, $\{Candidates\}$ denotes the text set predicted, and $\{Candidates\}$ denotes the text set of the reference.
                       \item \textbf{BERT SCORE}.To evaluate the semantically similarity between the candidate sentence and its reference sentence, BERT SCORE is defined based on the contextual embedding for vector representations for the word depending on its surrounding words, which is follows:
   \begin{eqnarray}
&\hspace{-0.4cm} BERT-P=\frac{1}{\mid y \mid}\sum\limits_{y_{j}\in y}\max\limits_{x_{i}\in x}\mathbf{x}_{i}^{T}\mathbf{y}_{j} \notag
\\
 &\hspace{-0.4cm} BERT-R=\frac{1}{\mid x \mid}\sum\limits_{x_{i}\in x}\max\limits_{y_{j}\in y}\boldsymbol{x}_{i}^{T}\boldsymbol{y}_{j} \notag \\
&\hspace{-0.4cm}  BERT-F= \frac{2BERT-R\times BERT-P}{BERT-P+BERT-R}
\end{eqnarray}
    where $x=\langle x_{1}, \cdots, x_{m} \rangle$,$\langle \mathbf{x}_{1}, \cdots, \mathbf{x}_{m} \rangle$, $y=\langle y_{1}, \cdots, y_{n} \rangle$,$\langle \mathbf{y}_{1}, \cdots, \mathbf{y}_{n} \rangle$ denote tokenized reference sentence, sequence of embedding vectors for the tokenized reference sentence, tokenized candidate sentence, sequence of embedding vectors for the tokenized candidate sentence, respectively. 
   \end{itemize}
\subsection{Evaluation of Text Regression}
For the prediction of prison term at a continuous interval, we can take them as text regression problems, such as in~\cite{2019Charge}.
Given the difference between the predicted prison term $\hat{y}_{i}$ and the ground truth value ${y}_{i}$  of the the $i$-th case, the evaluation metrics employed to evaluate the performance of prison term prediction problems are summarized as follows:
\begin{itemize}
    \item \textbf{Log distance}. To evaluate the tiny difference between the predicted prison term and its ground truth value based on distance, Log distance $LD$ is defined as follows:
        \begin{equation}
LD=\sum_{i=1}^{N}\frac{|log({y}_{i}+1)-log(\hat{y}_{i}+1)|}{N}
\end{equation}
     \item \textbf{S score}. As the metrics used in the CAIL2018 Competition~\cite{zhong2018overview}, $S$ score metric can be obtained to evaluate the similarity between two 
continuous stochastic variables :  
     
     \begin{equation}
S=\sum_{i=1}^{N}\frac{f(|log({y}_{i}+1)-log(\hat{y}_{i}+1)|)}{N}
\end{equation}

where, function $f(\cdot)$ satisfies:
\begin{equation}
f(v)=\left\{
\begin{array}{rcl}
1.0, & & {if \quad v \leq 0.2,}\\
0.8, & & {if \quad 0.2 < v \leq 0.4,}\\
0.6, & & {if \quad 0.4 < v \leq 0.6,}\\
0.4, & & {if \quad 0.6 < v \leq 0.8,}\\
0.2, & & {if \quad 0.8 < v \leq 1.0,}\\
0.0, & & {if \quad 1.0 < v. }
\end{array}
 \right.
\end{equation}
    \end{itemize}

\section{Methods}\label{methods}
 We introduce the methods for legal judgment prediction based on the publicly available datasets described in Section \ref{Datasets}.
If the LJP dataset consists of multiple subtasks, the multi-task learning (MTL) method can be used to utilize the dependencies among prediction results of multiple subtasks (see Section \ref{Multi-Task method}). If the dataset is a task-specific dataset, Pre-trained language models (PTMs, Section \ref{Pre-trained Language method}) can be used for fine-tuning downstream tasks. The few-shot learning (FSL) method (Section~\ref{Few-Shot learning method}) can be used for a few-shot scenario that the distribution of the dataset's judgment results is imbalanced.
\subsection{Multi-Task Learning}\label{Multi-Task method}
Multi-task learning (MTL) has numerous successful usages in NLP tasks, which transfers useful information across relevant tasks by solving them simultaneously so that it has been applied to a wide range of areas, including NLP especially the legal domain. 

A family of multi-task learning methods focusing on utilizing logical dependencies among relative tasks is described as follows:

For example, to improve performance for predicting charges in the civil law system (e. g. France, Germany, Japan, and China) according to the useful information of relevant articles, FLA~\cite{2017Learning} proposed a two-stack attention-based neural network, dubbed Fact-Law Attention (FLA), to jointly model the charge prediction task and the relevant article extraction task in a unified framework. Specifically, one stack for fact embedding based on a sentence-level and a document Bi-directional Gated Recurrent Units (Bi-GRU), and the other for article embedding, dynamically generated for each case according to the fact-side clues as extra guidance.

Then, a topological multi-task learning framework model, TOPJUDGE\footnote{https://github.com/thunlp/TopJudge}~\cite{2018Legal}, has been proposed based on the facts of a case and topological dependencies among articles, charges, and prison terms. Specifically, in the civil law system, given the fact description of a specific case, applicable law articles are firstly decided by a judge, and then the charges according to the instructions of relevant law articles are determined. Based on these results, the judge further confirms the terms of the penalty.

Furthermore, a Multi-Perspective Bi-Feedback Network (MPBFN) with the Word Collocation Attention (WCA) mechanism has been proposed in~\cite{2019Legal}. Specifically, the semantic vector of fact with word collocation and number semantic attention mechanism and the judgment results of pre-dependent tasks are utilized to make the forward prediction for its follow-up tasks. Meanwhile, the judgment results of follow-up tasks are also utilized to make a backward verification to check the rationality of their pre-order tasks.

Another family of multi-task learning methods is also widely used since they distinguish confusing follow-up tasks utilized the discriminative information of pre-order tasks. For example, a discriminative confusing charges model is proposed by incorporating relevant essential attributes (Few-Shot charge prediction model\footnote{https://github.com/thunlp/attribute\_charge}~\cite{2018few-shot}) or by distinguishing confusing law articles (Law Article Distillation based Attention Network, LADAN\footnote{https://github.
com/prometheusXN/LADAN
}~\cite{2020Distinguish}). 
\subsection{Pre-trained Language Model}\label{Pre-trained Language method}
Transformer-based~\cite{vaswani2017attention} pre-trained language models (PLMs), such as BERT~\cite{devlin-etal-2019-bert} and its variants (\cite{liu2019roberta,yang2019xlnet,beltagy2020longformer,chang2020dart,he2020deberta,zaheer2020big,sanh2019distilbert,lewis2019bart,JMLR:v21:20-074,xue2020mt5,liu2020multilingual,brown2020language}), have achieved
state-of-the-art results in several downstream NLP
tasks on generic benchmark datasets. As shown in Table \ref{table7}, inspired by the success of PLMs in the generic domain, there have been several approaches for applying BERT-based models in legal domain pretraining to explore state-of-the-art performance in downstream legal tasks. 

For example, LEGAL-BERT\footnote{https://huggingface.co/nlpaueb.}~\cite{2020LEGAL}, a new family of BERT models with 12 GB of English legal training corpora, consists of LEGAL-BERT-FP version (adapting standard BERT by additional pretraining on legal domain corpora) and LEGAL-BERT-SC version ( pretraining BERT from scratch on legal domain corpora). CaseLaw-BERT~\cite{zhengguha2021} is also the LEGAL-BERT-SC model but uses the case law corpus and custom domain-specific vocabulary. LegalDB is a DistillBERT model pre-trained on English legal-specific training corpora too. Lawformer~\cite{xiao2021lawformer} is a Longfomer-based model pre-trained on large-scale Chinese legal long case documents. JuriBERT~\cite{2021Douka} is a new set of BERT models, consisting of LEGAL-BERT-SC model pre-trained on French legal text datasets and adapting CamemBERT by additional pretraining on French legal text datasets.

Besides the effort to the domain-specific pre-trained language models, an improvement of PLMs on legal tasks with a document length of longer than 512 has been another research pot in the legal domain. For example, a hierarchical version of BERT, HIER-BERT~\cite{2020LEGAL}, concatenates BERT-BASE with a hierarchical attention network to bypass BERT’s length limitation.
Lawformer~\cite{xiao2021lawformer}, a Longformer-based pre-trained language model, combines the local sliding window attention and the global task motivated full attention to capturing the long-distance dependency for processing the Chinese legal documents with thousands of tokens.
\subsection{Interpretable-Perspective Learning Framework}
Interpretability in LJP, which means the ability of LJP systems to explain their predictions, has drawn increasing attention rapidly from academia and the legal industry, as people cannot wholly trust the machine-generated judgment results without any interpretation provided. The concept of interpretation~\cite{2016Generating} has been categorized into introspection explanation and justification explanation. Introspection explains how a model determines its final output. Justification explanation produces sentences detailing how the evidence is compatible with the system output.

Reinforcement learning methods are widely used for introspection explanation in LJP. Rationales from input fact description, where rationales serve as the introspection explanation for the charge prediction, are extracted using a deep reinforcement learning method~\cite{2018Interpretable}. QAjudge\footnote{https://github.com/thunlp/QAjudge}~\cite{2020Iteratively} is based on reinforcement learning to interpret the judgments. Specifically, a question net will select questions from the given set, and an answer net will answer the question according to the fact description. Finally, a predicted net will produce judgment results based on the answers, and reward functions are designed to minimize the number of questions asked.

Moreover, Multi-task learning methods in Section~\ref{Multi-Task method} can also be used for an introspection explanation of LJP.

Correspondence-based methods among relative tasks can also be used as an introspection explanation of LJP. For example, a fine-grained fact-article correspondence method is proposed~\cite{2021Learning} for recommending relevant law articles to a given legal case since the existing recommended articles do not explain which specific fact each article is relevant. Charge-based prison term prediction (CPTP)~\cite{2019Charge} has been proposed to make the total prison term prediction more interpretable based on the fine-grained charge-prison term correspondence feature selection and aggregation.

From the justification explanation aspect, Court Views~\cite{Hai2018Interpretable} has been considered as the explanation for the prediction of charges.

\subsection{Few-Shot Learning Framework}\label{Few-Shot learning method}
Few-shot learning has become a hot spot~\cite{zhao2017gated,2018few-shot,xu2020data,chang2021training,chang2021selectgen} since current works focus more on high-frequency judgment results than few-shot judgment results to ensure enough training data. 

Several few-shot learning methods can be used to predict few-shot judgment results. 

For example, discriminative attributes of charges~\cite{2018few-shot} are utilized to provide additional information for few-shot charges. A Sequence Enhanced Capsule model, dubbed the SECaps model~\cite{2018SECaps}, is based on the focal loss to predict few-shot charges. Moreover, an attentional and Counterfactual based Natural Language Generation (AC-NLG) method~\cite{2018SECaps} has been proposed, where the counterfactual decoder is used to address the imbalance problem in judgments.
\begin{table}
\centering
\caption{The current results for ECHR-CASES.}
\label{table10}
\begin{tabular}{|c|cc|}
\hline
\multirow{3}{*}{Method}       & \multicolumn{2}{c|}{Articles}                              \\ \cline{2-3} 
                              & \multicolumn{1}{c|}{Binary}            & Multi-lable       \\ \cline{2-3} 
                              & \multicolumn{1}{c|}{MaF}               & MaF               \\ \hline
BOW-SVM                       & \multicolumn{1}{c|}{70.9±0.0}          & 50.4±0.0          \\ \hline
BIGRU-ATT                     & \multicolumn{1}{c|}{78.9±1.9}          & 56.2±1.3          \\ \hline
HAN                           & \multicolumn{1}{c|}{80.2±2.7}          & 59.9±0.5          \\ \hline
HIER-BERT                     & \multicolumn{1}{c|}{\textbf{80.1±1.1}} & \textbf{60.0±1.3} \\ \hline
LEGAL-BERT-FP 100k ALL LEGAL  & \multicolumn{1}{c|}{\textbf{88.3}}     & 63.9              \\ \hline
LEGAL-BERT-FP 500k ALL LEGAL  & \multicolumn{1}{c|}{88.0}                & 60.3              \\ \hline
LEGAL-BERT-FP 100k SUB-DOMAIN & \multicolumn{1}{c|}{87.9}              & 60.5              \\ \hline
LEGAL-BERT-FP 500k SUB-DOMAIN & \multicolumn{1}{c|}{88.0}                & \textbf{65.2}     \\ \hline
\end{tabular}
\end{table}
\section{Results \& Observations}\label{result}
\begin{table*}
\centering
\caption{The current results for CAIL-Long dataset.}
\label{table11}
\begin{tabular}{|c|ccccc|cc|}
\hline
\multirow{3}{*}{Method} & \multicolumn{5}{c|}{Criminal}                                                                                                                                    & \multicolumn{2}{c|}{Civil}                     \\ \cline{2-8} 
                        & \multicolumn{2}{c|}{Charges}                                            & \multicolumn{2}{c|}{Articles}                                         & Prison terms   & \multicolumn{2}{c|}{Articles}                  \\ \cline{2-8} 
                        & \multicolumn{1}{c|}{MiF}           & \multicolumn{1}{c|}{MaF}           & \multicolumn{1}{c|}{MiF}         & \multicolumn{1}{c|}{MaF}           & LD             & \multicolumn{1}{c|}{MiF}         & MaF         \\ \hline
BERT                    & \multicolumn{1}{c|}{94.800}          & \multicolumn{1}{c|}{68.200}          & \multicolumn{1}{c|}{81.500}        & \multicolumn{1}{c|}{52.9}          & 1.286          & \multicolumn{1}{c|}{61.700}        & 31.600        \\ \hline
RoBERTa                 & \multicolumn{1}{c|}{94.700}          & \multicolumn{1}{c|}{69.300}          & \multicolumn{1}{c|}{81.100}        & \multicolumn{1}{c|}{53.500}          & 1.291          & \multicolumn{1}{c|}{60.200}        & 29.900        \\ \hline
L-RoBERTa               & \multicolumn{1}{c|}{94.900}          & \multicolumn{1}{c|}{70.800}          & \multicolumn{1}{c|}{81.100}        & \multicolumn{1}{c|}{53.400}          & 1.280           & \multicolumn{1}{c|}{61.200}        & 31.300        \\ \hline
\textbf{Lawformer}      & \multicolumn{1}{c|}{\textbf{95.400}} & \multicolumn{1}{c|}{\textbf{72.100}} & \multicolumn{1}{c|}{\textbf{82.000}} & \multicolumn{1}{c|}{\textbf{54.300}} & \textbf{1.264} & \multicolumn{1}{c|}{\textbf{63.000}} & \textbf{33.000} \\ \hline
\end{tabular}
\end{table*}
\begin{table*}
\centering
\caption{The current results for CAIL-2018 dataset.}
\label{table9}
\resizebox{\textwidth}{1in}{
\begin{tabular}{|c|ccc|ccc|ccc|}
\hline
\multirow{2}{*}{Method} & \multicolumn{3}{c|}{Acc}                                                                & \multicolumn{3}{c|}{MaP}                                                            & \multicolumn{3}{c|}{MaR}                                                            \\ \cline{2-10} 
                        & \multicolumn{1}{c|}{Charges}        & \multicolumn{1}{c|}{Articles}       & Prison terms   & \multicolumn{1}{c|}{Charges}        & \multicolumn{1}{c|}{Articles}       & Prison terms   & \multicolumn{1}{c|}{Charges}        & \multicolumn{1}{c|}{Articles}       & Prison terms   \\ \hline
FLA+MTL                 & \multicolumn{1}{c|}{92.76}          & \multicolumn{1}{c|}{93.23}          & 57.63          & \multicolumn{1}{c|}{76.35}          & \multicolumn{1}{c|}{72.78}          & 48.93          & \multicolumn{1}{c|}{68.48}          & \multicolumn{1}{c|}{64.30}           & 45.00             \\ \hline
CNN+MTL                 & \multicolumn{1}{c|}{95.74}          & \multicolumn{1}{c|}{95.84}          & 55.43          & \multicolumn{1}{c|}{86.49}          & \multicolumn{1}{c|}{83.20}           & 45.13          & \multicolumn{1}{c|}{79.00}             & \multicolumn{1}{c|}{75.31}          & 38.85          \\ \hline
HARNN+MTL               & \multicolumn{1}{c|}{95.58}          & \multicolumn{1}{c|}{95.63}          & 57.38          & \multicolumn{1}{c|}{85.59}          & \multicolumn{1}{c|}{81.48}          & 43.50           & \multicolumn{1}{c|}{79.55}          & \multicolumn{1}{c|}{74.57}          & 40.79          \\ \hline
Few-Shot+MTL            & \multicolumn{1}{c|}{96.04}          & \multicolumn{1}{c|}{96.12}          & 57.84          & \multicolumn{1}{c|}{88.30}           & \multicolumn{1}{c|}{85.43}          & 47.27          & \multicolumn{1}{c|}{80.46}          & \multicolumn{1}{c|}{80.07}          & 42.55          \\ \hline
TOPJUDGE                & \multicolumn{1}{c|}{95.78}          & \multicolumn{1}{c|}{95.85}          & 57.34          & \multicolumn{1}{c|}{86.46}          & \multicolumn{1}{c|}{84.84}          & 47.32          & \multicolumn{1}{c|}{78.51}          & \multicolumn{1}{c|}{74.53}          & 42.77          \\ \hline
MPBFN-WCA               & \multicolumn{1}{c|}{95.98}          & \multicolumn{1}{c|}{96.06}          & 58.14          & \multicolumn{1}{c|}{\textbf{89.16}} & \multicolumn{1}{c|}{85.25}          & 45.86          & \multicolumn{1}{c|}{79.73}          & \multicolumn{1}{c|}{74.82}          & 39.07          \\ \hline
LADAN+MTL               & \multicolumn{1}{c|}{96.45}          & \multicolumn{1}{c|}{96.57}          & 59.66          & \multicolumn{1}{c|}{88.51}          & \multicolumn{1}{c|}{86.22}          & \textbf{51.78} & \multicolumn{1}{c|}{\textbf{83.73}} & \multicolumn{1}{c|}{\textbf{80.78}} & 45.34          \\ \hline
LADAN+TOPJUDGE          & \multicolumn{1}{c|}{96.39}          & \multicolumn{1}{c|}{\textbf{96.62}} & 59.70           & \multicolumn{1}{c|}{88.49}          & \multicolumn{1}{c|}{\textbf{86.53}} & 51.06          & \multicolumn{1}{c|}{82.28}          & \multicolumn{1}{c|}{79.08}          & 45.46          \\ \hline
LADAN+MPBFN             & \multicolumn{1}{c|}{\textbf{96.42}} & \multicolumn{1}{c|}{96.60}           & \textbf{59.85} & \multicolumn{1}{c|}{88.45}          & \multicolumn{1}{c|}{86.42}          & 51.75          & \multicolumn{1}{c|}{83.08}          & \multicolumn{1}{c|}{80.37}          & \textbf{45.59} \\ \hline
\end{tabular}
}
\end{table*}
This section compares empirical results on LJP datasets via various NLP models. We focused on datasets from 8 sources: European Court of Human Rights (ECHR), Supreme People's Court of China (SPC), French Supreme Court (FSC), UK's highest Court (UKC), the Supreme Court of the United States (SCOTUS), French Supreme Court (FSC), (UKC)Supreme Court of Indian (SPC), Thai Supreme Court (TSC), and Federal Supreme Court of Switzerland (FSCS). We introduced 31 LJP datasets and six unlabelled pre-trained corpora above, but we focus on nine datasets from different source Court cases for looking into the results of experiments on them. Others have a limited size or limited numbers of experiments than the focused datasets.
\subsection{ECHR cases}
\begin{table*}
\centering
\caption{The current results for FSC, UKC, SCOTUS and SCI cases.}
\label{FUS cases}
\begin{tabular}{|c|c|c|cccc|}
\hline
\multirow{2}{*}{Dataset}                                                     & \multirow{2}{*}{Source} & \multirow{2}{*}{Method} & \multicolumn{4}{c|}{Plea judgments}                                                                                              \\ \cline{4-7} 
                                                                             &                         &                         & \multicolumn{1}{c|}{Acc}            & \multicolumn{1}{c|}{MaF}            & \multicolumn{1}{c|}{MaP}            & MaR            \\ \hline
Sulea et al.~\cite{sulea2017predicting}                     & FSC                     & \textbf{SVM}            & \multicolumn{1}{c|}{\textbf{96.90}}  & \multicolumn{1}{c|}{\textbf{97.00}}    & \multicolumn{1}{c|}{\textbf{97.10}}  & \textbf{96.90}  \\ \hline
\multirow{4}{*}{BStricks\_LDC~\cite{strickson2020legal}} & \multirow{4}{*}{UKC}    & \textbf{TFIDF+LR}       & \multicolumn{1}{c|}{\textbf{69.05}} & \multicolumn{1}{c|}{\textbf{69.02}} & \multicolumn{1}{c|}{\textbf{69.05}} & \textbf{69.02} \\ \cline{3-7} 
                                                                             &                         & LDA+kNN                 & \multicolumn{1}{c|}{57.76}          & \multicolumn{1}{c|}{57.63}          & \multicolumn{1}{c|}{58.01}          & 57.88          \\ \cline{3-7} 
                                                                             &                         & Word2Vec+RF             & \multicolumn{1}{c|}{64.21}          & \multicolumn{1}{c|}{64.17}          & \multicolumn{1}{c|}{64.17}          & 64.18          \\ \cline{3-7} 
                                                                             &                         & Count+RF                & \multicolumn{1}{c|}{66.13}          & \multicolumn{1}{c|}{66.12}          & \multicolumn{1}{c|}{66.12}          & 66.13          \\ \hline
\multirow{7}{*}{JUSTICE~\cite{2021JUSTICE2021}}              & \multirow{7}{*}{SCUS}   & Perceptron              & \multicolumn{1}{c|}{65.00}          & \multicolumn{1}{c|}{65.00}          & \multicolumn{1}{c|}{65.00}          & 65.00          \\ \cline{3-7} 
                                                                             &                         & SVM                     & \multicolumn{1}{c|}{60.00}          & \multicolumn{1}{c|}{60.00}          & \multicolumn{1}{c|}{60.00}          & 60.00          \\ \cline{3-7} 
                                                                             &                         & LR                      & \multicolumn{1}{c|}{61.00}          & \multicolumn{1}{c|}{61.00}          & \multicolumn{1}{c|}{61.00}          & 61.00          \\ \cline{3-7} 
                                                                             &                         & Naive Bayes             & \multicolumn{1}{c|}{59.00}          & \multicolumn{1}{c|}{59.00}          & \multicolumn{1}{c|}{59.00}          & 59.00          \\ \cline{3-7} 
                                                                             &                         & MLP                     & \multicolumn{1}{c|}{64.00}          & \multicolumn{1}{c|}{64.00}          & \multicolumn{1}{c|}{64.00}          & 64.00          \\ \cline{3-7} 
                                                                             &                         & \textbf{kNN}            & \multicolumn{1}{c|}{\textbf{68.00}} & \multicolumn{1}{c|}{\textbf{67.00}} & \multicolumn{1}{c|}{\textbf{69.00}} & \textbf{68.00} \\ \cline{3-7} 
                                                                             &                         & Clalib. Claasifier      & \multicolumn{1}{c|}{62.00}          & \multicolumn{1}{c|}{62.00}          & \multicolumn{1}{c|}{63.00}          & 62.00          \\ \hline
\multirow{5}{*}{ILDC~\cite{malik2021ildc}}                  & \multirow{5}{*}{SCI}    & Doc2Vec+LR              & \multicolumn{1}{c|}{60.91}          & \multicolumn{1}{c|}{62.00}          & \multicolumn{1}{c|}{63.03}          & 61.00          \\ \cline{3-7} 
                                                                             &                         & GloVe+BiGRU+att.        & \multicolumn{1}{c|}{60.75}          & \multicolumn{1}{c|}{64.35}          & \multicolumn{1}{c|}{68.26}          & 60.87          \\ \cline{3-7} 
                                                                             &                         & RoBERTa                 & \multicolumn{1}{c|}{71.26}          & \multicolumn{1}{c|}{71.77}          & \multicolumn{1}{c|}{72.25}          & 71.31          \\ \cline{3-7} 
                                                                             &                         & \textbf{XLNet+BiGRU}    & \multicolumn{1}{c|}{\textbf{77.78}} & \multicolumn{1}{c|}{\textbf{77.79}} & \multicolumn{1}{c|}{\textbf{77.80}} & \textbf{77.78} \\ \cline{3-7} 
                                                                             &                         & XLNet+BiGRU+att.        & \multicolumn{1}{c|}{77.01±0.52}   & \multicolumn{1}{c|}{77.07±0.01} & \multicolumn{1}{c|}{77.32}          & 76.82          \\ \hline
\end{tabular}
\end{table*}
Table \ref{table10} shows the evaluation results of recent studies on ECHR-CASES. BOW-SVM is a frequently used classification baseline model using Support Vector Machine (SVM)  based on the bag-of-words features. BIGRU-ATT is the standard sequence model Bi-directional Gated Recurrent Units (Bi-GRU,~\cite{bahdanau2014neural}) with attention, and Hierarchical Attention Network (HAN, ~\cite{yang2016hierarchical}) is also a sequential model.  HIER-BERT is a Hierarchical Transformer model to bypass BERT's length limitation, and LEGAL-BERT-FP 100k/500k ALL LEGAL/SUB-DOMAIN are transformer models with running additional pre-training steps (e.g., up to 100k or 500k) of BERT-BASE on legal-domain corpora (such as all legal corpora or just ECHR-CASES). As the tendency, LEGAL-BERT-FP 100k ALL LEGAL widely outperforms the other methods in binary violation classification, and LEGAL-BERT-FP 500k SUB-DOMAIN and LEGAL-BERT-FP 100k ALL LEGAL perform excellently. These results indicate that one can consider further pre-training to port BERT to a new domain.
\subsection{SPC cases}
\subsubsection{CAIL-Long Dataset}
Table \ref{table11} shows evaluation results of recent studies on CAIL-Long. Both BERT (~\cite{devlin-etal-2019-bert}) and RoBERTa (~\cite{liu1907roberta,cui2019pre}) are mainstream pre-trained language models (PLMs) with length limitations in the generic domain. Both L-RoBERTa and Lawformer are further pre-trained on the same legal corpus. Nevertheless, Lawformer is a Longformer-based PLMs on legal-domain corpora to process long documents. As the tendency, both in criminal and civil cases, Lawformer based on capturing the long-distance dependency for processing the long legal documents achieves the best performance, which indicates that one can consider Lawformer as the pre-trained model to finetune a legal document in Chinese longer than 512.
\subsubsection{CAIL-2018 Dataset}
Table \ref{table9} shows the evaluation results of recent studies on
CAIL2018. For the detailed explanations of methods, see
Section V. As the tendency, LADAN based on automatically extracting discriminative features of confusing articles
from fact descriptions achieve best accuracy Acc, macro-precision (MaP), and macro-recall (MaR).
MPBFN-WCA method shows better performance than the TOPJUDGE model due to bi-directional dependencies among LJP subtasks being utilized other than forwarding dependencies.
Both the MPBFN-WCA method and the TOPJUDGE model perform poorly under the MP, MR, and F1 metrics, which
indicates their shortage of predicting few-shot charges. In
addition, the performance of Few-Shot on charge prediction
is close to LADAN, but worse performance on prison term
due to the limitation of predefined attributes to help to predict
prison term.

\subsection{FSC, UKC, SCOTUS, and SCI cases}
\begin{table}
\centering
\caption{The evaluation results for fact snippets vs experts in ILDC dataset.}
\label{ILDC}
\begin{tabular}{|c|cccccc|}
\hline
\multirow{2}{*}{Expert} & \multicolumn{6}{c|}{Fact snippets vs Expets}                                                                                                                                                              \\ \cline{2-7} 
                        & \multicolumn{1}{c|}{JS}             & \multicolumn{1}{c|}{R-1}            & \multicolumn{1}{c|}{R-2}            & \multicolumn{1}{c|}{R-L}            & \multicolumn{1}{c|}{BLEU}          & Meteor       \\ \hline
\textbf{Expert 1}       & \multicolumn{1}{c|}{\textbf{0.333}} & \multicolumn{1}{c|}{0.444}          & \multicolumn{1}{c|}{\textbf{0.303}} & \multicolumn{1}{c|}{0.439}          & \multicolumn{1}{c|}{0.160}          & 0.220         \\ \hline
\textbf{Expert 2}       & \multicolumn{1}{c|}{0.317}          & \multicolumn{1}{c|}{\textbf{0.517}} & \multicolumn{1}{c|}{0.295}          & \multicolumn{1}{c|}{0.407}          & \multicolumn{1}{c|}{\textbf{0.280}} & \textbf{0.300} \\ \hline
Expert 3                & \multicolumn{1}{c|}{0.328}          & \multicolumn{1}{c|}{0.401}          & \multicolumn{1}{c|}{0.296}          & \multicolumn{1}{c|}{0.423}          & \multicolumn{1}{c|}{0.099}         & 0.180         \\ \hline
\textbf{Expert 4}       & \multicolumn{1}{c|}{0.324}          & \multicolumn{1}{c|}{0.391}          & \multicolumn{1}{c|}{0.297}          & \multicolumn{1}{c|}{\textbf{0.444}} & \multicolumn{1}{c|}{0.093}         & 0.177        \\ \hline
Expert 5                & \multicolumn{1}{c|}{0.318}          & \multicolumn{1}{c|}{0.501}          & \multicolumn{1}{c|}{0.294}          & \multicolumn{1}{c|}{0.407}          & \multicolumn{1}{c|}{0.248}         & 0.279        \\ \hline
\end{tabular}
\end{table}
\begin{table*}
\centering
\caption{The current results for FSCS cases.}
\label{FSCS cases}
\begin{tabular}{|c|cccccc|}
\hline
\multirow{3}{*}{Method}           & \multicolumn{6}{c|}{Plea judgments}                                                                                                                                                                                \\ \cline{2-7} 
                                  & \multicolumn{2}{c|}{de}                                                     & \multicolumn{2}{c|}{fr}                                                     & \multicolumn{2}{c|}{it}                                \\ \cline{2-7} 
                                  & \multicolumn{1}{c|}{MiF}           & \multicolumn{1}{c|}{MaF}               & \multicolumn{1}{c|}{MiF}           & \multicolumn{1}{c|}{MaF}               & \multicolumn{1}{c|}{MiF}           & MaF               \\ \hline
\textbf{Majority}                 & \multicolumn{1}{c|}{\textbf{80.3}} & \multicolumn{1}{c|}{44.5}              & \multicolumn{1}{c|}{\textbf{81.5}} & \multicolumn{1}{c|}{44.9}              & \multicolumn{1}{c|}{\textbf{81.3}} & 44.8              \\ \hline
Stratified                        & \multicolumn{1}{c|}{66.7±0.3}      & \multicolumn{1}{c|}{50.0±0.4}          & \multicolumn{1}{c|}{66.3±0.2}      & \multicolumn{1}{c|}{50.0±0.4}          & \multicolumn{1}{c|}{69.9±1.8}      & 48.8±2.4          \\ \hline
Linear(BoW)                       & \multicolumn{1}{c|}{65.4±0.2}      & \multicolumn{1}{c|}{52.6±0.1}          & \multicolumn{1}{c|}{71.2±0.1}      & \multicolumn{1}{c|}{56.6±0.2}          & \multicolumn{1}{c|}{67.4±0.5}      & 53.9±0.6          \\ \hline
Native BERT                       & \multicolumn{1}{c|}{74.0±4.0}      & \multicolumn{1}{c|}{63.7±1.7}          & \multicolumn{1}{c|}{74.7±1.8}      & \multicolumn{1}{c|}{58.6±0.9}          & \multicolumn{1}{c|}{76.1±3.7}      & 55.2±3.7          \\ \hline
Multilingual BERT                 & \multicolumn{1}{c|}{68.4±5.1}      & \multicolumn{1}{c|}{58.2±4.8}          & \multicolumn{1}{c|}{71.3±4.3}      & \multicolumn{1}{c|}{55.0±0.8}          & \multicolumn{1}{c|}{77.6±2.4}      & 53.0±1.1          \\ \hline
\textbf{long Native BERT}         & \multicolumn{1}{c|}{76.5±3.7}      & \multicolumn{1}{c|}{67.9±1.8}          & \multicolumn{1}{c|}{77.2±3.4}      & \multicolumn{1}{c|}{68.0±1.8}          & \multicolumn{1}{c|}{77.1±3.9}      & \textbf{59.8±4.6} \\ \hline
long Multilingual BERT            & \multicolumn{1}{c|}{75.9±1.6}      & \multicolumn{1}{c|}{66.5±0.8}          & \multicolumn{1}{c|}{73.3±1.9}      & \multicolumn{1}{c|}{64.3±1.5}          & \multicolumn{1}{c|}{76.0±2.6}      & 58.4±3.5          \\ \hline
\textbf{hierarchical Native BERT} & \multicolumn{1}{c|}{77.1±3.7}      & \multicolumn{1}{c|}{\textbf{68.5±1.6}} & \multicolumn{1}{c|}{80.2±2.0}      & \multicolumn{1}{c|}{\textbf{70.2±1.1}} & \multicolumn{1}{c|}{75.8±3.5}      & 57.1±6.1          \\ \hline
hierarchical Multilingual BERT    & \multicolumn{1}{c|}{76.8±3.2}      & \multicolumn{1}{c|}{57.1±0.8}          & \multicolumn{1}{c|}{76.3±4.1}      & \multicolumn{1}{c|}{67.2±2.9}          & \multicolumn{1}{c|}{72.4±16.6}     & 55.5±9.5          \\ \hline
\end{tabular}
\end{table*}
\begin{table}
\centering
\caption{The current results for TSC cases.}
\label{TSC cases}
\begin{tabular}{|c|cc|}
\hline
\multirow{2}{*}{Method} & \multicolumn{2}{c|}{Plea judgments}                   \\ \cline{2-3} 
                        & \multicolumn{1}{c|}{MiF}             & MaF            \\ \hline
Naive Bayes             & \multicolumn{1}{c|}{57.32}           & 57.28          \\ \hline
SVM                     & \multicolumn{1}{c|}{64.23}           & 60.45          \\ \hline
BiGRU                   & \multicolumn{1}{c|}{58.54}           & 57.99          \\ \hline
BiGRU + self-att.       & \multicolumn{1}{c|}{59.76}           & 59.11          \\ \hline
BiGRU + att.            & \multicolumn{1}{c|}{\textbf{66.68}} & \textbf{63.35} \\ \hline
\end{tabular}
\end{table}
Table \ref{FUS cases} shows the evaluation results of recent studies on FSC, UKC, SCOTUS, and SCI cases. All Support Vector Machine (SVM), Logistic Regression (LR), K-Nearest Neighbor (kNN), Naive Bayes, Perceptron, multi-layer perceptron (MLP), calibrated classifier (Calib. Classifier), and Random Forest are classical feature-based machine learning models (e.g., using word/sentence embedding and TFIDF). RoBERTa and XLNet are mainstream transformer models. As the tendency, XLNet with BiGRU outperforms classical and sequential models on the ILDC dataset, the best performing classical model on UKC cases is TFIDF with LR on the top, kNN is the top-performing classical model on SCOTUS cases, and the classical model performs excellently on FSC cases. These results show that transformer-based sequence models may further improve the performance of classical models. In addition, Table \ref{ILDC} shows exact matching results between automatically extracted Fact snippets by machine and annotated references by experts on the ILDC dataset, indicating that an efficient explainable model for the LJP task may point a direction of future research. 

\subsection{FSCS cases}
Table \ref{FSCS cases} shows the evaluation results of recent studies on FSCS cases. All Majority, Stratified and Linear (BoW) are baseline classifiers, where Majority selects the majority class across all the legal cases, Stratified predicts the judgment labels randomly, and Linear (BoW) is a linear classifier using TFIDF features. Long BERT introduces additional positional embedding to extend the limited sequence length of the standard BERT model. Moreover, hierarchical BERT separates and encodes the input tokens with a standard BERT encoder and then aggregates all segment encodings with an additional Bidirectional Long Short-Term Memory (BiLSTM) to form the representation of classification. As the tendency, the majority method performs best compared to other baseline models and BERT-based models in MiF. Considering the error margin, both the long BERT model and hierarchical BERT (German and French with 20K+ training samples) outperform the Italian ones with 3K training samples in MaF. These results indicate that imbalanced class distribution, the scale of a dataset, and the variants of BERT models may influence the performance.
\subsection{TSC cases}
Table \ref{TSC cases} shows the evaluation results of recent studies on TSC cases. SVM and Naive Bayes are frequently used classification baseline models. Both of these two models are non-neural models. As the tendency, sequence model BiGRU with attention outperforms the non-neural models on the TSCC dataset.

To summarize, experiments on ECHR-CASES, CAIL-Long, ILDC, and FSCS cases indicate that further pre-trained BERT-based models on legal corpora could predict the judgment better than using classical/sequence models or that with BERT-based models on generic corpora. And, experiments on CAIL-2018 show that the combination of two families of MTL outperforms other models on solving multi-task simultaneously. And evaluation results on the ILDC dataset also indicate that interpretable learning for the LJP task is still challenging work. In addition, experiments on FSCS cases show that BERT-based methods without employing few-shot learning on imbalanced label distribution can underperform that on balanced label distribution.
\section{Discussions \& Recommendations}\label{Discussions}
This section analyzes different LJP datasets and solutions and discusses our findings.

\begin{itemize}
    \item Sophisticated LJP Datasets from real courtrooms. As the performances on LJP are improved, doing more fine-grained and sophisticated real court judgment will become of more practical significance. Up to now, a large number of large-scale judge-summarized facts description-based LJP datasets, such as CAIL2018 or CAIL-Long in Table~\ref{table00}, have been built. However, these datasets could threaten the case logic representation quality and prediction correctness since they neglect the effect of the admissibility of evidence and facts in a real court setting. For all we know, the LJP dataset from real courtrooms is only LJP-MSJudge~\cite{RN202}, but this dataset only considers ten facts labels of 70,482 Private Lending cases. Hence, we propose the recommendations for a new dataset stating the actual legitimate intent, especially when the rules of proof are inadequate and the text of evidence is doubt or ambiguous.
    \item Complex legal logical reasoning. An ideal legal judgment predictor is expected to examine the case from different views~\cite{RN202} comprehensively. As shown in Table~\ref{table9}, improving the predicting accuracy of prison terms is highly urgent, although the accuracy of prediction for articles and charges is over 90\%. How to augment legal knowledge from various views and machine reasoning in NLP is of great significance.
    \item Adaptive interpretability. As shown in Table~\ref{FUS cases}, a pre-trained language model, XLNet with BiGRU, which outperforms classical and sequential models~\cite{malik2021ildc}, is the best performing model on the ILDC dataset. However, considering that the logic of practice judgments has its specialty for each law domain, the explanation for its black-box work will still be challenging in LJP tasks.
   \end{itemize}

\section{Conclusion}
In this survey, we first reveal the importance and history
of legal judgment prediction. Then we compare and
discuss the most recent benchmark datasets and experimental results of different methods. Based on our observations,
we propose new recommendations for future datasets and
also give the following suggestions for our future law judgment prediction model: predict whether the admissibility of evidence, complex reasoning features, and discriminative task-specific features can be incorporated into LJP datasets and existing neural network model for better performance, respectively. 

\section*{Acknowledgment}
This work was supported in part by the Science Foundation of The China(Xi'an) Institute for Silk Road Research (2019YB05, and 2019YA07), the Innovation training project of Shaanxi college students(S202011560030), and in part by the Research Foundation of Xi'an University of Finance and Economics under Grant 
18FCJH02.

\bibliographystyle{elsarticle-num}
\bibliography{cas-refs}
\end{document}